\documentclass[lettersize,journal]{IEEEtran}
\usepackage[switch]{lineno}
\usepackage{amsmath,amsfonts}
\usepackage{algorithmic}
\usepackage{algorithm}
\usepackage{array}
\usepackage{multirow}

\usepackage[caption=false,font=normalsize,labelfont=sf,textfont=sf]{subfig}
\usepackage{textcomp}
\usepackage{stfloats}
\usepackage{url}
\usepackage{verbatim}
\usepackage{graphicx}
\usepackage{cite}
\usepackage{epsfig}
\usepackage{orcidlink}

\begin{document}
\title{Towards Understanding the Benefits of Neural Network Parameterizations in Geophysical Inversions: A Study With Neural Fields}

\author{Anran Xu\,\orcidlink{0009-0001-9731-377X}, Lindsey J. Heagy\orcidlink{0000-0002-1551-5926}
\thanks{Manuscript received March 17, 2025.}
\thanks{Anran Xu and Lindsey J. Heagy are with the Department of Earth, Ocean and Atmospheric Sciences, University of British Columbia, Vancouver, BC V6T 1Z4 Canada (email: anranxu@eoas.ubc.ca; lheagy@eoas.ubc.ca)}}

\maketitle

\begin{abstract}
Recent research in test-time machine learning methods has shown that some machine learning models, without any prior learning, can improve the results of geophysical inversions. Some examples include the Deep Image Prior Inversions (DIP-Inv) and the Neural Fields Inversions (NFs-Inv), where the inverse problems are reparametrized by the weights of the machine learning models. In this work, we employ neural fields, which use neural networks to map a coordinate to the corresponding physical property value at that coordinate, in a test-time learning manner. For a test-time learning method, the weights are learned during the inversion, as compared to traditional approaches, which require a network to be trained using a training dataset. Results for synthetic examples in seismic tomography and direct current resistivity inversions are shown first. We then perform a singular value decomposition analysis on the Jacobian of the weights of the neural network (SVD analysis) for both cases to explore the effects of neural networks on the recovered model. The results show that the test-time learning approach can eliminate unwanted artifacts in the recovered subsurface physical property model caused by the sensitivity of the survey and physics. Therefore, NFs-Inv improves the inversion results compared to the conventional inversion in some cases, such as the recovery of the dip angle or the prediction of the boundaries of the main target. In the SVD analysis, we observe similar patterns in the left-singular vectors as were observed in some diffusion models, trained in a supervised manner, for generative tasks in computer vision. This observation provides evidence that there is an implicit bias, which is inherent in the neural network structures, that is useful in supervised learning and test-time learning models. This implicit bias has the potential to be useful for recovering models in geophysical inversions.

\end{abstract}

\begin{IEEEkeywords}
Deep neural networks (DNN), neural fields (NFs), deep learning (DL), test-time learning (TTL), implicit bias, inductive bias, direct-current resistivity (DCR) inversion, and cross-hole seismic tomography inversion.
\end{IEEEkeywords}

\section{Introduction}
\label{sec:1}
\IEEEPARstart{G}{eophysical} inversion is an important step for interpreting observed geophysical measurements \cite{menke_1989}, \cite{ref47}, \cite{cowan_airborne_2025}. In an inversion, a physical property model of the subsurface is estimated from measurements obtained on the surface, in boreholes or from airborne systems. The quality of the inversion results depends on robust forward modelling, reliable prior information, and well-designed regularization techniques. Due to the inherently ill-posed nature of geophysical inverse problems, developing a suitable regularization is challenging. Conventional regularization typically penalizes spatial variations and/or explicitly promotes proximity to a reference model \cite{fournier_inversion_2019}, \cite{Soler2025-kh}, but with advances in machine learning, there are opportunities to explore more advanced regularization.

Recently, machine learning methods have been demonstrated to be successful in many geophysical applications \cite{https://doi.org/10.1029/2021JB022581}, \cite{alyousuf_threeaxis_2024}, \cite{zhang_velocitygan_2019}, \cite{8994191}. Alyousuf \textit{et al.} \cite{alyousuf_threeaxis_2024} trained ML models to identify unexpected ${\rm CO_2}$ migration. Zhang \textit{et al.} \cite{zhang_velocitygan_2019} improved the reconstructions of subsurface seismic velocities by mapping velocity models to raw seismic waveforms using trained conditional adversarial networks. Liu \textit{et al.} \cite{8994191} predicted anomalous bodies in the subsurface by matching apparent resistivity data to the resistivity model using trained conventional neural networks.


Neural fields are advanced ML methods that can solve problems in 3D rendering, point cloud reconstruction, biomedical image recovery, etc \cite{https://doi.org/10.1111/cgf.14505}, \cite{10.1145/3503250}, \cite{9709943}, \cite{9606601}, \cite{Cachia_2023}. Neural fields use coordinate-based neural networks to map spatial, temporal, or other coordinates to the corresponding quantity at those coordinates. Neural fields are also called coordinate-based signal representations, or neural implicit representations, in certain contexts. Unlike the previously described data-driven methods, which map input-output pairs, neural fields use and view neural networks as a continuous representation of the target's properties in the coordinate space. Many applications of neural fields are not data-driven, instead, they are physics-informed or test-time learning. 

A significant and successful application of neural fields is in 3D rendering, which aims to generate new views of complex scenes using a set of input views \cite{10.1145/3503250}. The output values (colors and opacity) of the neural fields are fed into a forward map, which is based on a volume rendering algorithm, to generate the predicted observations (pixel values). The objective function, which measures the difference between the predicted measurements and real measurements, is used to update the weights of the neural network so that the loss value is expected to decrease iteratively. This training process is very similar to the process of solving inverse problems, so neural fields have been adapted to many inverse problems since then. The benefits of utilizing neural fields come from the following factors: (1) There is no need for pre-training, namely, the performance of these methods is not limited by the quality and quantity of the training dataset. (2) In principle, neural fields are mesh-free methods, namely, we can employ an adaptive grid scheme during inversion, which may bring benefits in terms of the accuracy of the reconstruction and the computational costs. (3) Parameterizing the inverse problems in a continuous setting naturally introduces smoothing regularization effects. (4) We can leverage the implicit/inductive regularization/bias from the neural network's structure and optimization method.

The performance of neural fields methods in 3D rendering has been shown to be much better than that of conventional methods in terms of distinguishing the boundaries between objects, predicting the pixel values of the main objects, etc \cite{10.1145/3503250}. The sources of the benefits of utilizing the neural fields are still controversial. For example, Yu \textit{et al.} \cite{yu_plenoxels_2021} argue that the neural network is not the source of these benefits by showing that when the neural network in the NeRF, a typical NFs method in 3D rendering, is replaced with spherical harmonics, the performance in the 3D rendering tasks is not affected. Many other works do not explicitly discuss the sources of benefits by employing the neural fields methods in 3D rendering \cite{10.1145/3503250}, \cite{martinbrualla2021nerfwildneuralradiance}. A key consideration when assessing and designing a model is its performance in representation tasks. For example, in the context of 3D rendering, researchers want to find a model that can reconstruct the 2D image better by learning self-similarity. The main strategies for improving model performance include: selecting a positional encoding function tailored to the problem \cite{tancik2020fourierfeaturesletnetworks}, \cite{DBLP}, improving the activation function \cite{sitzmann2020implicitneuralrepresentationsperiodic}, and employing different bases \cite{fathony2021multiplicative}, \cite{yu_plenoxels_2021}. Notably, the majority of the works still utilize neural networks as a component in the neural field methods.

Neural fields methods have also garnered significant attention from researchers outside of the domain of computer vision and computer graphics. A main domain for testing the usage of neural fields methods is biomedical imaging \cite{arratia2024enhancingdynamicctimage}, \cite{molaei2023implicitneuralrepresentationmedical}, \cite{9606601}, \cite{shen_nerp_2024}, \cite{liu2022recoverycontinuous3drefractive}. In CT and MRI imaging reconstruction tasks, Shen \textit{et al.} \cite{shen_nerp_2024} showed that learning the implicit neural representation of the reconstructed images by adapting a prior embedding can significantly improve the resolution of the reconstruction results. Liu \textit{et al.} \cite{liu2022recoverycontinuous3drefractive} used neural fields with a pre-trained denoising regularizer and showed that this approach could improve the recovery of 3D refractive index images from discrete measurements.  
Sun \textit{et al.} \cite{9606601} improve reconstruction performance in CT tomographic imaging with a coordinate-based neural network by generating new measurements from existing data and incorporating these new measurements during training. 

In the geosciences, Rasht-Behesht \textit{et al.} \cite{https://doi.org/10.1029/2021JB023120} proposed physics-informed neural networks for seismic inversions by parameterizing two physical quantities in the wave equation with two coordinate-based neural networks. The results show that this method can provide a good recovery of the subsurface model without the need for an explicit regularization term. Sun \textit{et al.} \cite{IFWI} tested the performance of parameterizing the seismic velocity directly with a coordinate-based neural network in the full-waveform inversion. Both works argue that neural fields methods can improve the reconstruction performance in scenarios lacking accurate starting models. Neural fields have also been applied to the joint inversion of seismic and electrical data. 
Liu \textit{et al.} \cite{Mingliang} used a Bayesian neural network to parameterize the global model properties and a convolutional neural network to parameterize the reservoir properties. 
The numerical results show that the proposed pipeline provides a more accurate recovery of the reservoir model. Notably, some advanced techniques in computer vision and computer graphics were adapted to some of the scientific inverse problems mentioned above and proved to be useful. For example, both Rasht-Behesht \textit{et al.} \cite{https://doi.org/10.1029/2021JB023120} and Sun \textit{et al.} \cite{IFWI} mentioned the importance of using an adaptive sinusoidal activation function. Liu \textit{et al.} \cite{liu2022recoverycontinuous3drefractive} showed the benefits of adopting a task-specific positional encoding function. There are some works in geophysical inverse problems on parameterizing the subsurface models using neural networks \cite{https://doi.org/10.1029/2025JH000621}, \cite{https://doi.org/10.1029/2024JH000542}, but there are only a limited number of works that use neural fields methods, and a majority of them focus on seismic inversions. Thus, there is substantial potential to further explore this new area.

There are many previous works on Physics-Informed Neural Networks (PINNs), which also use coordinate-based neural networks in solving scientific inverse problems. In general, PINNs incorporate governing physical laws, such as partial differential equations, directly into deep neural networks (DNNs) as prior knowledge. This approach addresses the challenge of data scarcity in training ML models to solve scientific inverse problems \cite{https://doi.org/10.1029/2020JB020549}, \cite{https://doi.org/10.1029/2019WR026731}, \cite{gmd-16-7375-2023}. Compared to the direct-feedforward networks, PINNs can achieve better accuracy and shorter training time by using a smaller training dataset \cite{RAISSI2019686}, \cite{cite-key}. Our work does not fall into the category of PINN mainly because we do not let the neural network learn the physics of the problem. Namely, we enforce the PDE constraints by using a PDE solver to calculate the residual instead of using a PDE residual where the PDE is directly embedded in the objective function, so there is no need for any training dataset.  

It is well-known that learning alone is insufficient to explain the good performance of deep networks in terms of generalizability. The fact that deep neural networks can describe the underlying distribution of a dataset, even in high-dimensional spaces that suffer from the curse of dimensionality, suggests that there is a prior (implicit/inductive bias) that has been imposed in the training process \cite{Zahra}. However, the underlying mechanism of this implicit bias has not been fully explained. For example, some researchers have found that certain modern ML models' structures (e.g. Transformers, CNNs, DNNs) have an implicit regularization or bias that is beneficial for some inverse problems \cite{22}, \cite{ref49}, \cite{10.1145/3571070}, \cite{frei2022implicitbiasleakyrelu}. Some of them suggested that the benefits come from the high dimensionality of the neural network \cite{frei2022implicitbiasleakyrelu}, \cite{allenzhu2019convergencetheorydeeplearning}, \cite{arora2018optimizationdeepnetworksimplicit}. Others analyze this generalizability using probabilistic theory \cite{wilson2022bayesiandeeplearningprobabilistic}, \cite{griffiths2023bayesageintelligentmachines}. Although some of these works have taken a mathematical point of view and have aimed to prove rigorous theorems, they have been limited to overly simplified settings. Other works used the empirical-based method (or the so-called ``scientific method"), which forms hypotheses and designs controlled experiments to test them \cite{Physics_of_LLMS}. Specifically, it highlights the importance of empirical analyses of deep neural networks to validate or refute existing theories and assumptions, as well as to address fundamental questions regarding the mechanisms behind their successes. Although relatively under-explored, these empirical approaches offer significant potential to strengthen our understanding of deep learning and to drive substantial advancements in both theoretical and practical domains. Some well-known works that emphasize empirical-based methods include: Physics of Large Language Models \cite{Physics_of_LLMS}, where Zhu \textit{et al.} used synthetic data and built an idealized environment for the training of LLMs to find the ``laws of all LLMs". A scaling law for LLMs states that the performance of LLMs scales with model size, dataset size, and the amount of compute power \cite{kaplan2020scalinglawsneurallanguage}. Kadkhodaie \textit{et al.} \cite{Zahra} found that the diffusion models tend to learn a basis that is adapted to geometric features in the underlying images, and they name these bases as “geometry-adaptive harmonic bases”. 

Kadkhodaie \textit{et al.} \cite{Zahra} attributed the implicit bias to the model structure and optimization method, not the training data alone. Therefore, we can utilize the analysis of that work to take one step further to examine the performance of the implicit bias and test-time learning methods in the geophysical inverse problems. Previous works have demonstrated the benefits of test-time learning methods to geophysical inversions and have attributed these benefits to the implicit bias provided by the machine learning model structures \cite{ref18}, \cite{27}. However, these works do not provide an analysis of what priors this implicit bias impose during the inversion. In this work, we will show that test-time machine learning methods can improve the geophysical inversion result by finding weights that can capture geometric structures in the physical property model. 

In this work, we will first introduce our NFs-Inv method and its applications in seismic tomography and direct current resistivity (DCR) inversions. We use the same structure of the Neural Network (NN) for both cases. However, we use different positional encoding functions in different cases since some positional encoding functions are helpful in terms of learning high-frequency variations of the subsurface model. The comparisons between the conventional Tikhonov-style inversion results and the NFs-Inv inversion results clearly show that our NFs-Inv method can improve the recovery of the subsurface model in some cases. We will then present our analysis of these inversion results. Our analysis, which uses a singular value decomposition (SVD) analysis on the Jacobian of the weights of the neural network, will partly explain why searching over the NN-weights space will improve the inversion results. To the best of our knowledge, this is the first study that applies neural fields methods to seismic tomography inversions and direct current resistivity inversions and gives evidence to show that the benefits of the test-time learning (TTL) method are from the implicit bias inherited from the NN's structure.

The main contributions of our work include: 
\begin{enumerate}
\item{Developing a methodology for using Neural Fields in geophysical inversions (NFs-Inv) in a test-time learning manner that does not require any pre-training or a training dataset. We also demonstrate the effect/influence of choosing different positional encodings in NFs-Inv.}
\item{Presenting its application on linear/non-linear geophysical inversions with homogeneous/inhomogeneous background. We compare the NFs-Inv and conventional inversion results. These results demonstrate that the implicit bias from the NFs-Inv is appropriate for some geophysical inversion scenarios.}
\item{Providing an empirically-based explanation of the benefits of employing NFs-Inv. The left-singular vectors of the Jacobian have a similar pattern to the ``geometry-adaptive harmonic bases" described by Kadkhodaie \textit{et al.} \cite{Zahra}. The basis of the Jacobian reveals what has been imposed by the implicit/inductive bias. The bias enables the inversion to overcome the problems of predicting unwanted artifacts due to sensitivity, such as near-electrode artifacts or the concentration along the ray path. This result suggests that the implicit bias in the NFs-Inv could be similar to the implicit bias found in certain dataset-trained machine learning models, such as diffusion models.}
\end{enumerate}



\section{Methodology}
\label{sec:2}
Many geophysical inversion methods can be categorized into a physics-driven approach or a data-driven approach. The physics-driven approach imposes explicit regularization to the objective function or uses a problem-specific constrained algorithm during the inversion process. The advantage is that it is a white-boxed approach--namely, the regularization and constraints have well-established geological, mathematical and/or statistical theories as a foundation. The physics-driven approach can be utilized when there is only limited knowledge of the survey domain, but it can sometimes be very time-consuming since it needs to run a forward simulation in each epoch during the test time. The data-driven approach typically requires more prior knowledge of the geological structures in the subsurface within the survey domain. This prior knowledge can be used to build a dataset of plausible subsurface geophysical models and then can be learned by the ML model trained using this dataset. The trained ML models can give fast predictions for new observations. This approach is black-boxed---namely, it is not guaranteed that the trained ML model can return a plausible subsurface model that fits the observations for any test data. Our work will combine elements of the two approaches: The proposed method is physics-informed since it includes a forward numerical simulation, which will make the method more robust and guarantee that the inversion results can fit the observations to a desired level. The proposed method also directly utilizes the implicit regularization from the coordinate-based representations without the need for a training dataset.

\subsection{Background and conventional inversion approach}

This study focuses on cross-hole seismic tomography inversion and 2D Direct Current (DC) resistivity inversion, which are widely used in mining, environmental, and engineering applications\cite{sharma_inversion_2015}, \cite{wei_quantifying_2022}, \cite{melo_geophysical_2017}. 

For cross-hole seismic tomography, the goal of the inversion is to recover the subsurface velocity model by analyzing the travel times of seismic waves between boreholes. In this survey, a seismic source is placed in one borehole, and seismic receivers (geophones) are placed in another borehole. The seismic source generates waves that travel through the subsurface and are detected by the receivers in the neighboring borehole. By moving the sources and receivers to various depths within their respective boreholes, multiple arrival times are recorded. The travel times of these waves are then used to create a seismic velocity model of the subsurface, which is related to the lithology and elastic properties of the subsurface. Using the straight-ray assumption, we can calculate the arrival time at each receiver for a given source using equation \ref{velocity equation}.
\begin{equation}
\label{velocity equation}
t_i = \int_{{\rm ray}_i} \frac{1}{v}(x,z) d\ell 
\end{equation}
where $t$ is the first arrival time and $v$ is the velocity of the seismic wave through the subsurface material. $\rm ray_i$ is the $i$th ray path from the given seismic source to a corresponding receiver.

In a typical DC resistivity survey, transmitters inject a steady-state electrical current into the ground, and receivers measure the resulting distribution of potentials (voltages). The governing equation for the DC resistivity experiment is
\begin{equation}
\label{Poisson equation}
\boldsymbol{\nabla} \cdot \sigma\boldsymbol{\nabla}\phi = -\boldsymbol{\nabla}\cdot\mathbf{j}_{source}
\end{equation}
where $\sigma$ is the electrical conductivity (which is the reciprocal of electrical resistivity $\rho=1/\sigma$), $\phi$ is the electric potential, and $\mathbf{j}_{source}$ is the source current density. We solve this equation using a finite volume method. For the examples we will show, we use a 2D tensor mesh. The mesh consists of a core region, with uniform cells, and padding regions on the lateral boundaries and the bottom of the mesh with expanding cells to ensure boundary conditions are satisfied. The forward modelling is implemented using SimPEG \cite{SimPEG}.  

In the conventional method, we use a tensor mesh for both surveys and denote the core domain as $m \in R^{W \times H}$. We initialize $m \in M$ as the reference model $m_{ref}$, a uniform half-space, and then iteratively update $m$ to reduce the output of the objective function. At each iteration, we input the current subsurface model $m$ into a forward numerical simulation $F$, whose output $F(m) = d^{pred}$ is then used to compute the data misfit term:
\begin{equation}
\begin{aligned}
\phi_{d}(m, d^{obs}) = \frac{1}{2}|W_d(d^{obs} - d^{pred})|^2.
\end{aligned}
\end{equation}
The objective function is a summation of the data misfit term $\phi_{d}(m, d^{obs})$ and the regularization term $\phi_{m}(m)$ with a trade-off parameter that weights the relative combination of each term. We assume that the noise is Gaussian and depends on the magnitude of the voltage measurements in the DCR survey. A matrix $W_d$, which includes data uncertainties, has been used to quantify the noise \cite{ref47}. For the cross-hole seismic tomography survey, we assume that the noise is Gaussian but is independent of the magnitude of the arrival time, so we adapt an identity matrix for $W_d$.

For the Tikhonov-style geophysical inversions, the regularization term $\phi_{m}(m)$ generally includes a smallness term and a smoothness term:
\begin{equation}
\label{conventinal}
\begin{aligned}
\phi_{m}(m) = \alpha_s|m-m_{ref}|^{p_s} + \alpha_x|\frac{\partial m}{\partial x}|^{p_x}+ \alpha_z|\frac{\partial m}{\partial z}|^{p_z}.
\end{aligned}
\end{equation}
Often, the norms applied are $p_{s,x,z}=2$, however, norms of less than 2 can also be employed to promote sparse or compact structures \cite{fournier_inversion_2019}. Regularization terms are commonly employed for the cases where we have no or limited geological prior knowledge. More advanced regularization can also be employed. For example, Astic and Oldenburg \cite{astic_framework_2019} proposed an inversion framework for integrating the petrophysical and geological prior into the direct current resistivity inversion. Laloy \textit{et al.} \cite{ref8} and Lopez-Alvis \textit{et al.} \cite{ref11} integrated the prior of distribution of the spatial patterns of the subsurface materials into the inverse process using a trained Variational Autoencoder. In this work, we will perform conventional inversions using the regularization in equation \ref{conventinal} as the benchmark. The hyperparameter values in equation \ref{conventinal} can be found in Section \ref{sec:3}.

\subsection{Proposed NFs-Inv Method}

\begin{figure*}[h!]
\centering
\includegraphics[width=7.in]{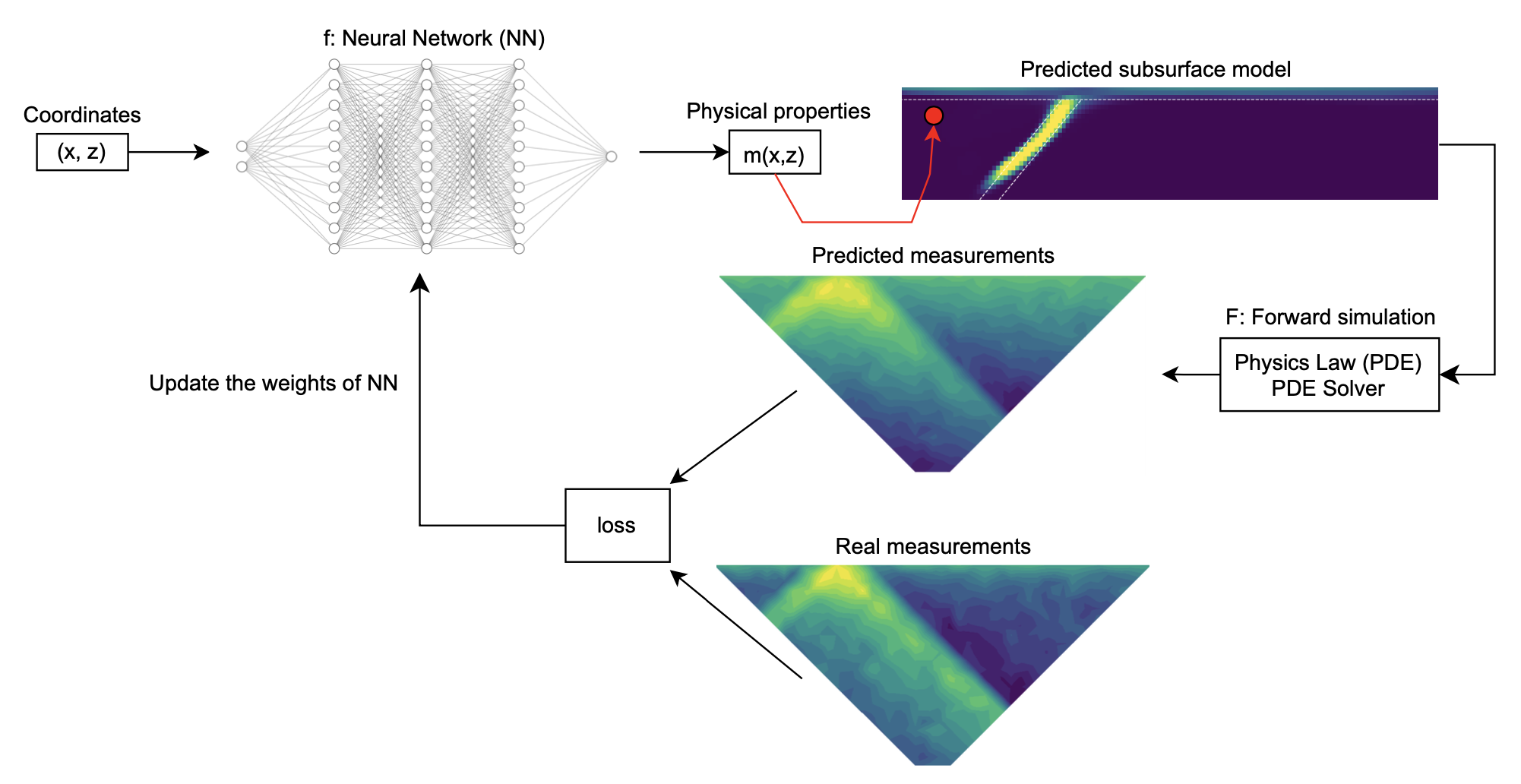}
\caption{Proposed inversion pipeline. We reparametrize the subsurface model $m$ by the weights of a NN. Considering a 2D DCR inversion, the input of NN f is the 2D spatial coordinates (x, y), and the output is the scalar value of the corresponding physical property (in this case, the conductivity value) at the coordinate. After obtaining physical properties on all coordinates, we have a 2D $m$ that is the predicted subsurface model at the current epoch. This 2D $m$ is then fed into the forward simulation to generate corresponding predicted measurements. After that, the loss function value is computed. During the inverse process, the weights of the NN are updated until the loss value is small enough.
}
\label{NF_scheme}
\end{figure*}

A schematic diagram of the proposed methods is shown in Fig. \ref{NF_scheme}. The key component of the proposed method is to reparameterize the model $m$ by the weights of a multilayer perceptron $f$. In the 2D setting, the input of $f$ is the 2D spatial coordinates of the mesh cell centers (x, z), and the output is the scalar value of the corresponding physical properties for which we invert (slowness or log-conductivity) on the coordinate.

Positional encoding is a transform function that maps (x, y) to a higher-dimensional space before feeding the input vector into the neural network to overcome spectral bias. Spectral bias is a well-known experimental finding stating that neural networks are biased towards learning lower-frequency variations \cite{rahaman2019spectralbiasneuralnetworks}. This is the reason why a direct mapping from the coordinate space to the physical property space using a neural network has poor performance in the image representation task---the high-frequency variation has not been learned at the time of convergence, so the reconstructed images are blurred and lack details \cite{rahaman2019spectralbiasneuralnetworks}. 
Positional encoding was first proposed in the work of NeRF \cite{10.1145/3503250} and then explained using the neural tangent kernel theory \cite{tancik2020fourierfeaturesletnetworks}. For context, we will introduce the positional encodings that are relevant to our work.
Let $\textbf{x} \in R^{w \times 1}$ denote a vector in the coordinate space, and the output of the encoding function is $\gamma(\textbf{x}) \in R^{h \times 1}$. The common positional encoding functions include \cite{10.1145/3503250}, \cite{9606601}, \cite{tancik2020fourierfeaturesletnetworks}, \cite{sun_coil_2021}:

\begin{flalign}
\label{Basic encoding}
\textbf{Basic encoding} \quad \quad \quad \quad \quad \quad \quad 
\gamma(\textbf{x}) = [cos(2\pi\textbf{x}), sin(2\pi\textbf{x})]^{T}.
\end{flalign}

\begin{equation}
\label{linear encoding}
\begin{split}
& \textbf{Linear positional encoding (firstly adopted by CoIL \cite{sun_coil_2021})} \\
& \gamma(\textbf{x}) = [cos(2\pi k^{1}\textbf{x}), sin(2\pi k^{1}\textbf{x}),...,cos(2\pi k^{m}\textbf{x}), sin(2\pi k^{m}\textbf{x})]^{T},\\
&\textrm{where} \quad k^{i} = \frac{i}{2}.
\end{split}
\end{equation}

\begin{equation}
\label{Gaussian encoding}
\begin{split}
& \textbf{Gaussian positional encoding} \\
& \gamma(\textbf{x}) = [cos(2\pi B\textbf{x}), sin(2\pi B\textbf{x})], \\
&\textrm{where B} \in R^{\hat{h} \times \hat{w}} (h = 2\hat{h}) \textrm{ is sampled from a Gaussian distribution}.
\end{split}
\end{equation}
Some other positional encodings, such as the radial Gaussian encoding \cite{liu2022recoverycontinuous3drefractive}, are not widely used, so we do not conduct experiments on them. The choice of positional encoding functions depends on the problems.

Let us denote this transform function as $\gamma: R^2 -> R^{h}$.
To obtain the subsurface model from our neural network, we require a $W \times H$ number of inputs, which is the same size as the number of active mesh cells in our inversion domain.
To simplify the computation, we construct a matrix $Z \in R^{W \times H, h}$ whose rows are the transformed spatial coordinates:
\begin{equation}
\begin{aligned}
Z_{ij} = [\gamma(x_i, y_i)]_j. 
\end{aligned}
\end{equation}
where $i$ is the index for the coordinates and $j$ is the index of the output of the transform function.
Then $W \times H$ can be considered as the batch size. Namely, the inversion results are gained by solving the optimization problem:

\begin{equation}
\label{proposed_objective}
\begin{aligned}
&\min_m \:\frac{1}{2}(1-\beta)\phi_{d}(m, d^{obs}) + \beta \phi_{m}(m)= \\
&\min_w \:\frac{1}{2}(1-\beta)\phi_{d}(f_{w}(Z), d^{obs}) + \beta \phi_{m}(f_{w}(Z)).
\end{aligned}
\end{equation}

There are several strategies for defining the trade-off parameter $\beta$. In many cases, a $\beta$-cooling strategy is adopted where the inversion starts with a large value for $\beta$ and $\beta$ is gradually decreased as the inversion progresses in order to fit the data. In the proposed method, we choose to use the same cooling schedule that is employed by the DIP-Inv method \cite{27}, where a test-time learning method has been applied to the DCR inversion. The trade-off parameter $\beta$ in the proposed method for the 2D DCR inversion is an exponential decay function shown in equation \ref{dacay_curve}.
\begin{equation}
\label{dacay_curve}
\beta = e^{-\frac{t}{\tau}},
\end{equation}
where t is the index for the epochs and $\tau$ is a constant decay rate. Namely, $\beta = e^{-\frac{i}{\tau}}$ in the i-th iteration. For the cross-hole seismic tomography inversion, we don't employ any explicit regularization term in the objective function in the proposed method, so $\beta = 0$ for these inversions.

The approach we take for solving this optimization problem is described in Algorithm \ref{alg:alg2}.
\begin{algorithm}
\caption{NFs-Inv algorithm.}
\label{alg:alg2}
\begin{algorithmic}
\STATE 
\STATE $\textbf{Input:}\;$ 
\STATE $\quad \beta$: the trade-off parameter for DCR inversion, 
\STATE $\quad Z$: encoded input of dimension $R^{W \times H, h}$, 
\STATE $\quad d^{obs}$: observed data (travel time, voltages, etc.), 
\STATE $\quad m_{ref}$: the reference subsurface model, \STATE $\quad f$: the multilayer perceptron,
\STATE $\quad w^t$: the initial weights of $f$, 
\STATE $\textbf{for}\; \textrm{t = 1}\; \textbf{to}\; \textrm{...k}\; \textbf{do}$
\STATE $\quad\textrm{Update the trade-off parameter}\; \beta^t$
\STATE $\quad\textrm{Predict subsurface model }\; m^t = f_{w^t} (Z)$
\STATE $\quad\textrm{Calculate the data misfit value}\; \phi_{d}(m^t, d^{obs})$
\STATE $\quad\textrm{Calculate the gradient w.r.t. the physical property}\; J_v = \nabla_{m^t} \phi_{d}(m^t, d^{obs})$
\STATE $\quad\textrm{Compute}\;\mathcal{L'} =(1-\beta^t)(J_v^T m^t) + \beta^t \phi_{m}(m^t)$
\STATE $\quad\textrm{Update}\; w_t \; \textrm{using } \mathcal{L'.}\textrm{backward() and optimizer.step()} *\;$
\STATE $\textrm{end for}\;$
\STATE $\textrm{Return final surface model $m^k$}\;$
\STATE $\textrm{*See DIP-Inv \cite{27} for explanation}\;$
\end{algorithmic}
\end{algorithm}

\subsection{Architecture of Neural Networks and Optimization}

The number of fully connected layers should be adapted to the complexity of the subsurface model and physics (e.g. it is a function of both the resolution of the forward problem and the complexity of the subsurface model). The input dimension of the first hidden layer should depend on the $\gamma$ function we choose. For example, if we choose the basic encoding \ref{Basic encoding} in 2D, then the input should have dimension 4. For both the cross-hole seismic tomography inversion and DCR inversion examples shown below, the results are obtained using a neural network, $f$, with six hidden layers with the LeakyReLU activation function. The first and last hidden layers have 128 neurons, and all other hidden layers have 256 neurons. For the DCR inversions, where we search the solution in the log-conductivity space, the activation function for the last layer is a Sigmoid function. For the cross-hole seismic tomography inversions, the activation function for the last layer is a Tanh function. The output in both cases is multiplied by a constant scalar to fit the possible ranges of slowness/conductivity. The details of choosing the $\gamma$ function will be discussed in Section \ref{sec:3}. 

Compared to the conventional method, the proposed method searches for a solution in a higher-dimensional space, so the number of iterations needed to fit the data is larger. The proposed method employs the Adam optimizer, which is a first-order optimization method. The conventional method in the cross-hole seismic tomography inversion employs a gradient descent optimizer, which is also a first-order optimization method. The conventional method in the direct current resistivity inversion employs the Inexact Gaussian Newton method, which is a second-order optimization method. The reason why we do not use a first-order optimization method in the direct current resistivity inversion is that the forward modelling, which is implemented using SimPEG, is much more time-consuming than the cross-hole seismic tomography inversion, so we want to adapt a second-order optimizer that can take fewer iterations to converge. We choose Adam for the proposed method since the adaptive learning rate scheme of Adam is beneficial when updating the weights in the neural networks \cite{kingma2017adammethodstochasticoptimization}. A second-order optimization method with an adaptive learning rate has been proposed but is limited by the computational cost of estimating the Hessian when the dimension of the parameter is high \cite{tan_lim_2019}.

\section{Trials on the Synthetic Examples}
\label{sec:3}
In this section, we will consider two geophysical inversion methods: the cross-hole seismic tomography inversion, a linear problem, and the direct current resistivity (DCR) inversion, a non-linear problem. We present the experiment results on 2D synthetic models to show the benefits of the proposed method. The proposed method can be readily adapted to 3D cases by changing the dimension of the input coordinate from two to three. In the cross-hole seismic tomography inversion, the data has sufficient resolution to recover a model even without a regularization, so we do not use an explicit regularization term in the proposed NF-Inv method. We will compare this result to the conventional method, both with and without an L2 smoothness term. In the direct current resistivity inversion, the data has less resolution, so in our proposed method, we use an L1 smallness term. For the conventional method, we will show results that utilize sparse norms on the smallness and smoothness and include a sensitivity weighting in the objective function.

\subsection{Trials on the Cross-hole Seismic Tomography Inversions}
\label{sec:3.1}
We will consider two models: the first is a simple model of a block in a uniform background, and the second is an elliptical target in a non-homogeneous background. For both cases, the tensor mesh has size $W \times H = 64 \times 128$, and each cell has a size of 1 meter by 1 meter. The two boreholes are separated by 64 meters. The receivers are evenly spaced on one borehole with a separation of 1 meter. The transmitters are evenly spaced on the other borehole with a separation of 1 meter.

In the forward simulation, we calculate arrival times (in units of milliseconds), and these are regarded as our geophysical observations. Gaussian noise with a mean of 0 ms and a standard deviation of 20 ms has been added to the observations for both cases before inversion. The PDE solver block in Fig. \ref{NF_scheme} refers to the linear forward modelling algorithm. 

\begin{figure}[h!]
\centering
\includegraphics[width=3.5in]{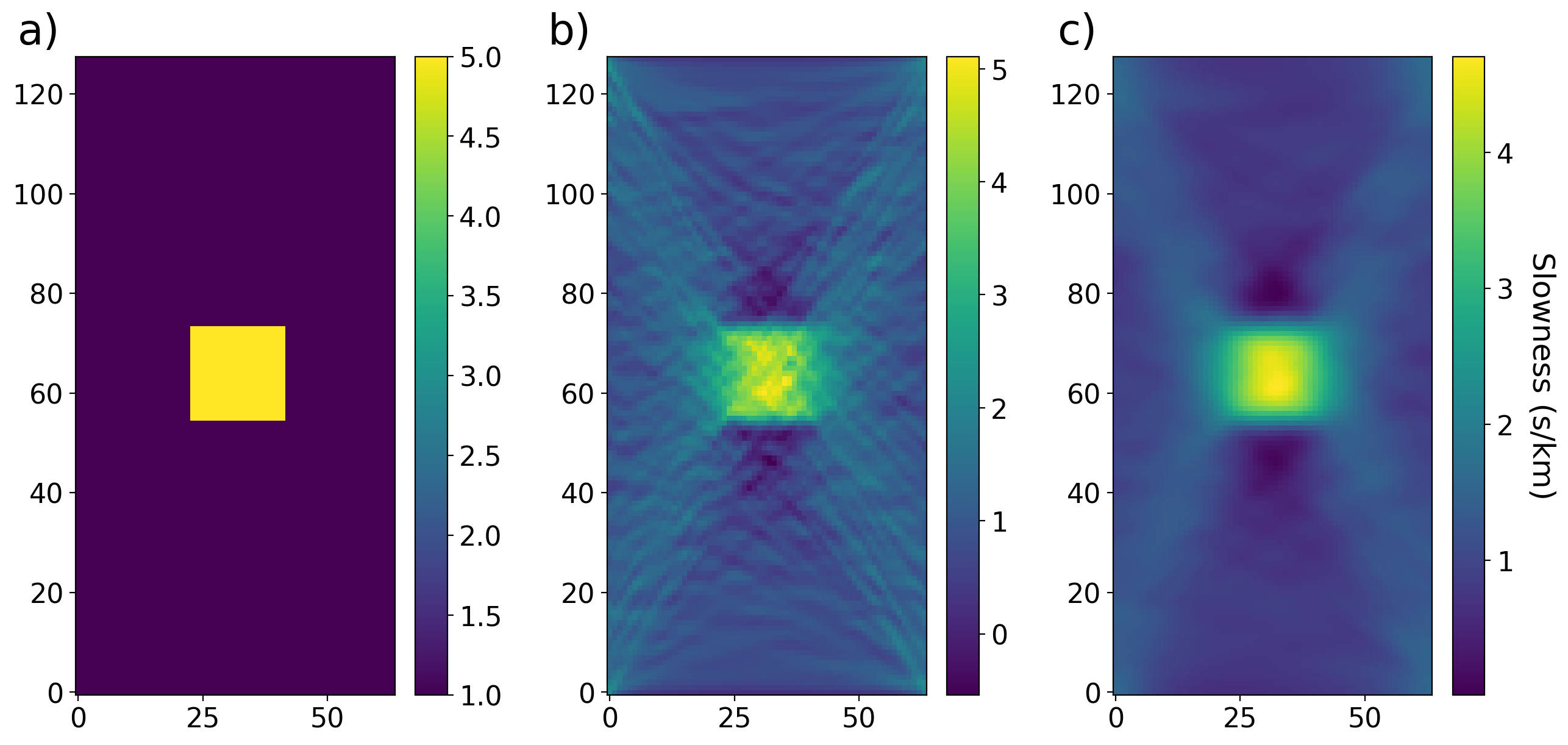}
\caption{The inversion results using the conventional method for Case 1. (a) is the true subsurface model. (b) is the conventional result without a smoothness term. (c) is the conventional result with the L2 smoothness term. The starting/initial model is a uniform half-space for the conventional methods.}
\label{NF_Case_1.1_L2}
\end{figure}

Fig. \ref{NF_Case_1.1_L2}(a) shows Case 1, a uniform background with a square target. The square target has a velocity of 200 m/s, and the background has a velocity of 1000 m/s. In the conventional method, the initial model is chosen to be a uniform background with a value of 1000 m/s. We perform two conventional inversions. Fig. \ref{NF_Case_1.1_L2}(b) shows the case where no explicit regularization was employed during the inverse process, and the recovered model tends to predict structures along the ray path. Fig. \ref{NF_Case_1.1_L2}(c) shows the conventional inversion result with only a smoothness term ($\alpha_s = 0$, $\alpha_x = \alpha_z = 0.5$, and $p_x = p_z = 2$). By adapting a smoothness term, the unwanted along-ray artifacts have been eliminated, but we can still see some along-ray-path artifacts remaining. 

\begin{figure}[h!]
\centering
\includegraphics[width=3.5in]{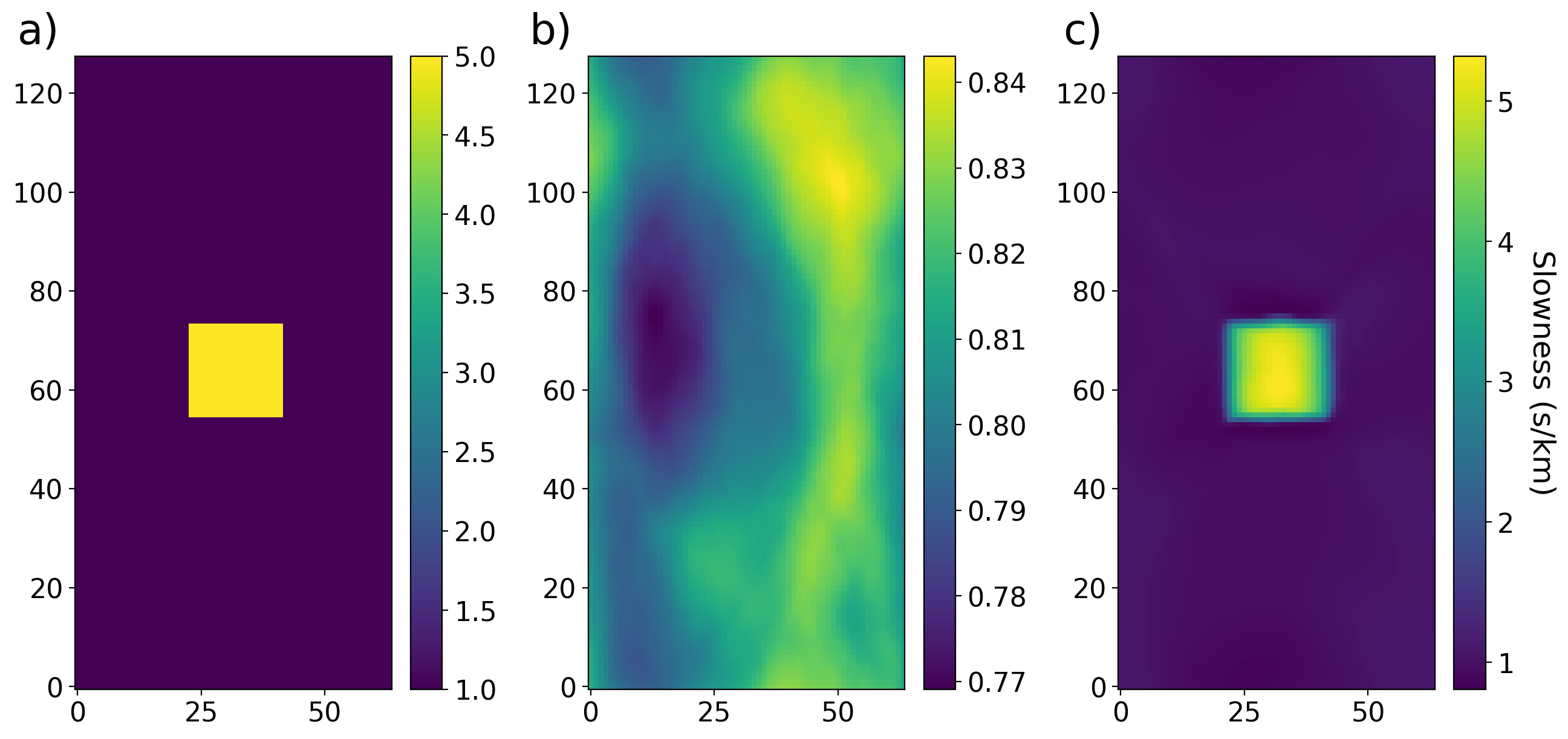}
\caption{The inversion results using the proposed NFs-Inv method for Case 1. (a) is the true subsurface model. (b) is the starting model. (c) is the recovered subsurface model from the proposed NFs-Inv method without any explicit regularization term.}
\label{NF_Case_1.1_new}
\end{figure}

In the proposed method, the input coordinate space is scaled to [0, 1] in both the x and z dimensions, and the initial model is randomly generated since the $m^1 = f_{w^1}(Z)$ and $w^1$ is randomly generated using the Kaiming Gaussian initialization method \cite{he2015delvingdeeprectifierssurpassing}. We choose the basic encoding, so the input dimension of $f$ is 4. For the Adam optimizer, the learning rate is 0.001. In Fig. \ref{NF_Case_1.1_new}, we show that the proposed method improves the reconstruction by eliminating the unwanted prediction along the ray path; therefore, the proposed inversion result provides a better recovery of the boundary of the main squared body.

\begin{figure}[h!]
\centering
\includegraphics[width=3.5in]{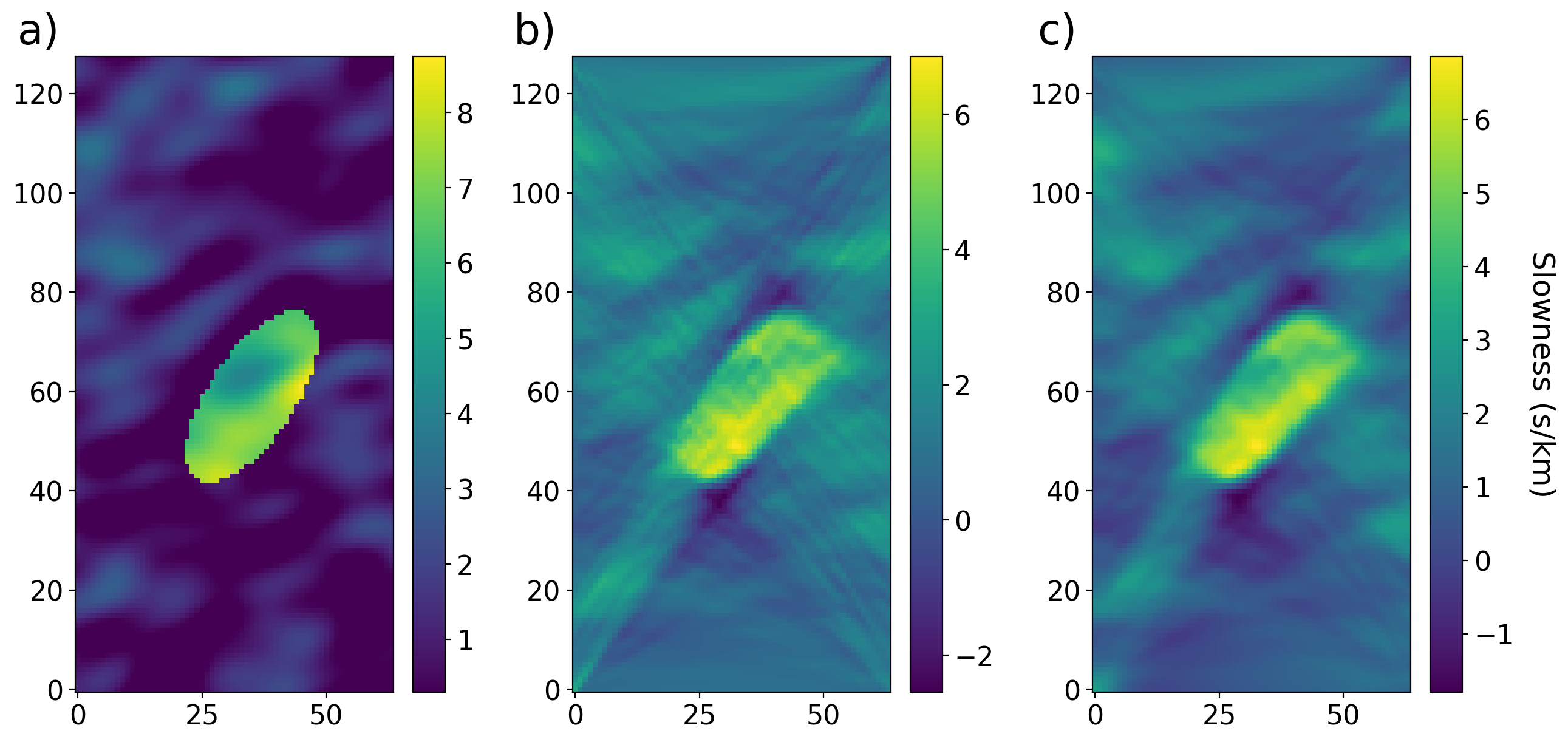}
\caption{The inversion results using the conventional method for Case 2. (a) is the true subsurface model. (b) is the conventional result without any regularization. (c) is the conventional result with the L2 smoothness term. The starting/initial model is a uniform half-space for the conventional methods.}
\label{NF_Case_2_L2}
\end{figure}

The second case we consider (Case 2) has a heterogeneous background, which was generated using the GeoStatTools package \cite{gmd-15-3161-2022}, with an elliptical target [Fig. \ref{NF_Case_2_L2}(a)]. The conventional method again has the uniform half-space as the initial model. 

\begin{figure}[h!]
\centering
\includegraphics[width=3.5in]{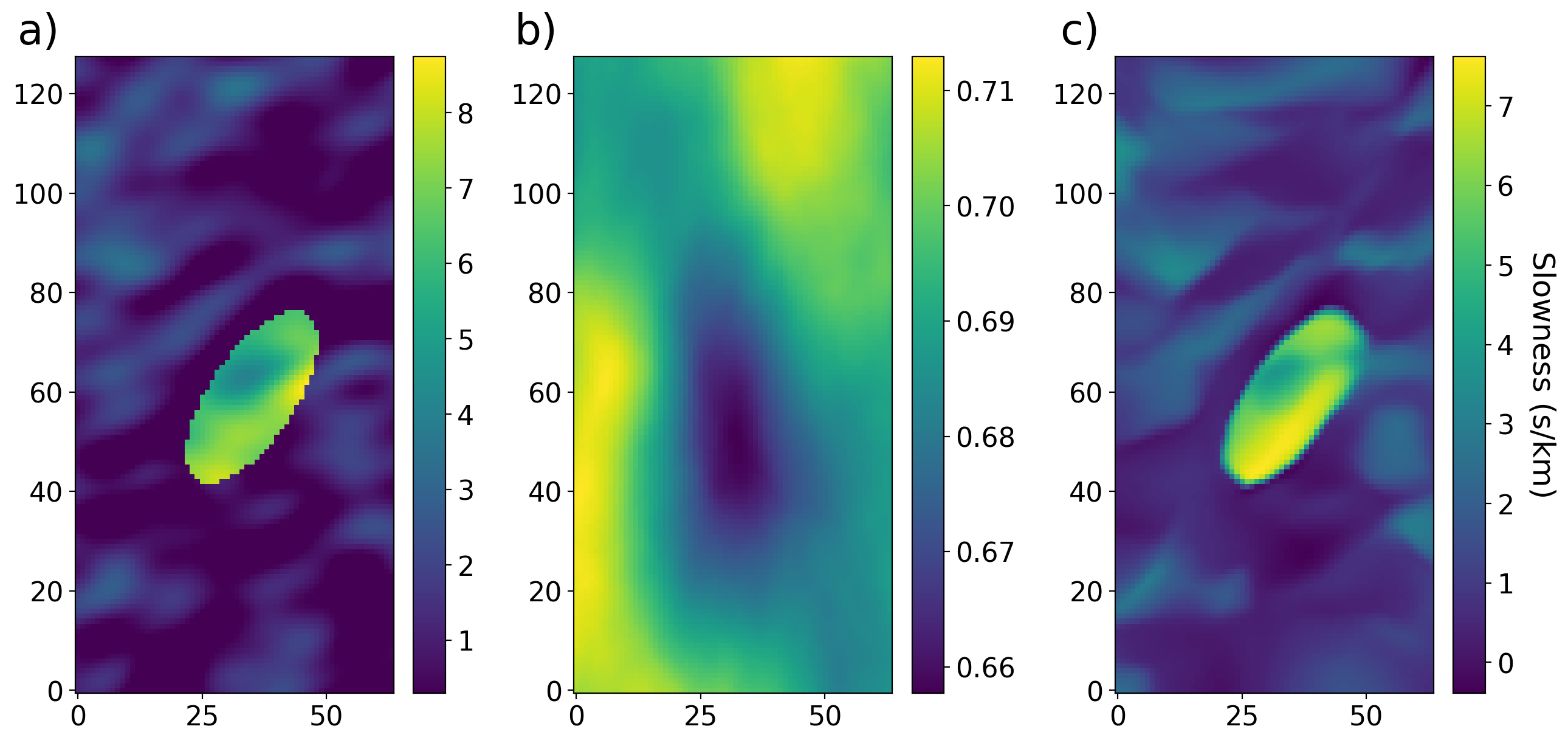}
\caption{The inversion results using the proposed NFs-Inv method for Case 2. (a) is the true subsurface model, (b) is the starting model, and (c) is the recovered subsurface model from the proposed NFs-Inv method without any explicit regularization term.}
\label{NF_Case_2_new}
\end{figure}

Similar to Case 1, we performed two inversions. Fig. \ref{NF_Case_2_L2}(b) shows the case where no explicit regularization was employed during the inverse process, and the recovered model tends to predict structures along the ray path. Due to the sensitivity and the lack of a positivity constraint, the recovered model has unrealistic negative values in some regions. Fig. \ref{NF_Case_2_L2}(c) shows the conventional inversion result with only a smoothness term ($\alpha_s = 0$, $\alpha_x = \alpha_z = 0.5$, and $p_x = p_z = 2$). By adding a smoothness term, the unwanted along-ray artifacts have been eliminated, but the edges of the target are blurred, and we can see some along-ray-path artifacts remaining.

In the proposed method, the input coordinate space is scaled to [-1, 1] in both the x and z dimensions, and the starting model is randomly initialized [Fig. \ref{NF_Case_2_new}(b)]. To have a better resolution on the inhomogeneous background where the high-frequency variations exist, we choose the Gaussian positional encoding with $B \in R^{128\times 2}$ sampled from a Gaussian distribution with a standard deviation 0.5, so the input dimension of $f$ is 256. The impact of the positional encoding will be discussed in Section \ref{sec:4.2}. For the Adam optimizer, the learning rate is 0.001. Compared to the recovered models in the conventional method [Fig. \ref{NF_Case_2_L2}(b), (c)], the NFs-Inv does not have any unwanted along-path structures. The heterogeneous background prediction has more details, and the boundaries of the main target are much better resolved [Fig. \ref{NF_Case_2_new}(c)]. Although we can have a better recovery by imposing a positive constraint (e.g. setting the activation function for the last layer to be ReLU), we employ the hyperbolic tangent as the activation function in the last layer to have a fair comparison with the conventional method where no positive constraint is imposed. Although there is no positive constraint in both methods, the proposed method performs better than the conventional method in terms of having fewer predictions of negative values.

\subsection{Trials on the Direct Current Resistivity Inversions}
\label{sec:3.2}

We again consider two cases (Cases 3 and 4). In both cases, there is a dipping target, but in Case 3, the background is homogeneous, whereas in Case 4, the resistivity of the background increases with depth. For both cases, we choose a dipole-dipole survey geometry with a station separation of 25 meters and a maximum of 24 receivers per transmitter. The survey line is 700 meters long, so the total number of potential difference measurements is 348. 

For the forward simulation, we use a mesh that has core cells of 5m $\times$ 5m, with 200 core-mesh cells in the x-direction and 45 cells in the z-direction. We add 7 padding cells that extend by a factor of 1.5 to the sides and bottom of the mesh. In total, the mesh has $W\times H = 9000$ active cells. In the proposed method, the input coordinate space is scaled to [-1, 1] in both the x and z dimensions in both cases. The input coordinate space only contains the core domain (active cells), and the padding domain values are set to be the background conductivity and frozen during the inversion process. We use the same positional encoding function (identity $\gamma$ function) and network structure for both cases, and the total number of trainable weights in the neural network in $f$ is 263809. 

\begin{figure*}[h!]
\centering
\includegraphics[width=7.in]{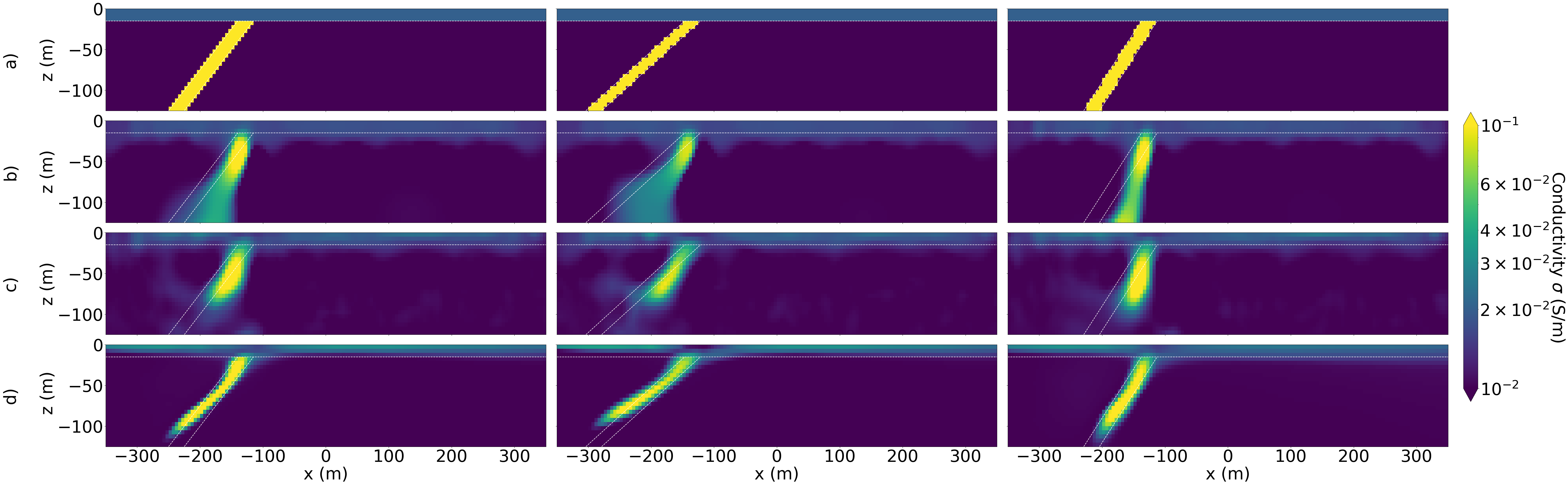}
\caption{The inversion results using the proposed method for Case 3. The top row (a) is the true model. The models in row (b) are from the conventional sparse-norm inversions with sensitivity weighting. The models in row (c) are from the DIP-Inv. The models in row (d) are from the proposed method.}
\label{NF_DIP_NF_compare}
\end{figure*}

For Case 3, the top layer has a conductivity of 0.02 S/m. The dike, with three different dip angles, has a conductivity of 0.1 S/m and extends to a depth of 125 meters [Fig. \ref{NF_DIP_NF_compare}(a)]. The background has a conductivity of 0.01 S/m. 
For the conventional method, we tried a range of parameter choices for the regularization. The best results we obtained used $\beta_s = 1e2$, $p_s = 0$, $p_x = p_z = 1$, $\alpha_s = 0.005$, and $\alpha_x = \alpha_z = 0.5$ [Fig. \ref{NF_DIP_NF_compare}(b)]. 

For the Adam optimizer, the learning rate is 0.001. For the $\beta$-cooling strategy in equation \ref{dacay_curve}, we choose $\tau = 800$. The results are shown in Fig. \ref{NF_DIP_NF_compare}(d). Considering another test-time learning method (DIP-Inv \cite{27}) has been developed and tested on the DCR inversion tasks, we also present the inversion result from DIP-Inv for comparison [Fig. \ref{NF_DIP_NF_compare}(c)].

Compared to the conventional methods, both DIP-Inv and the proposed method improve the dip angle predictions. Compared to the results of the DIP-Inv method, the predicted dike structures have sharper boundaries and thinner structures. 

\begin{figure}[h!]
\centering
\includegraphics[width=3.5in]{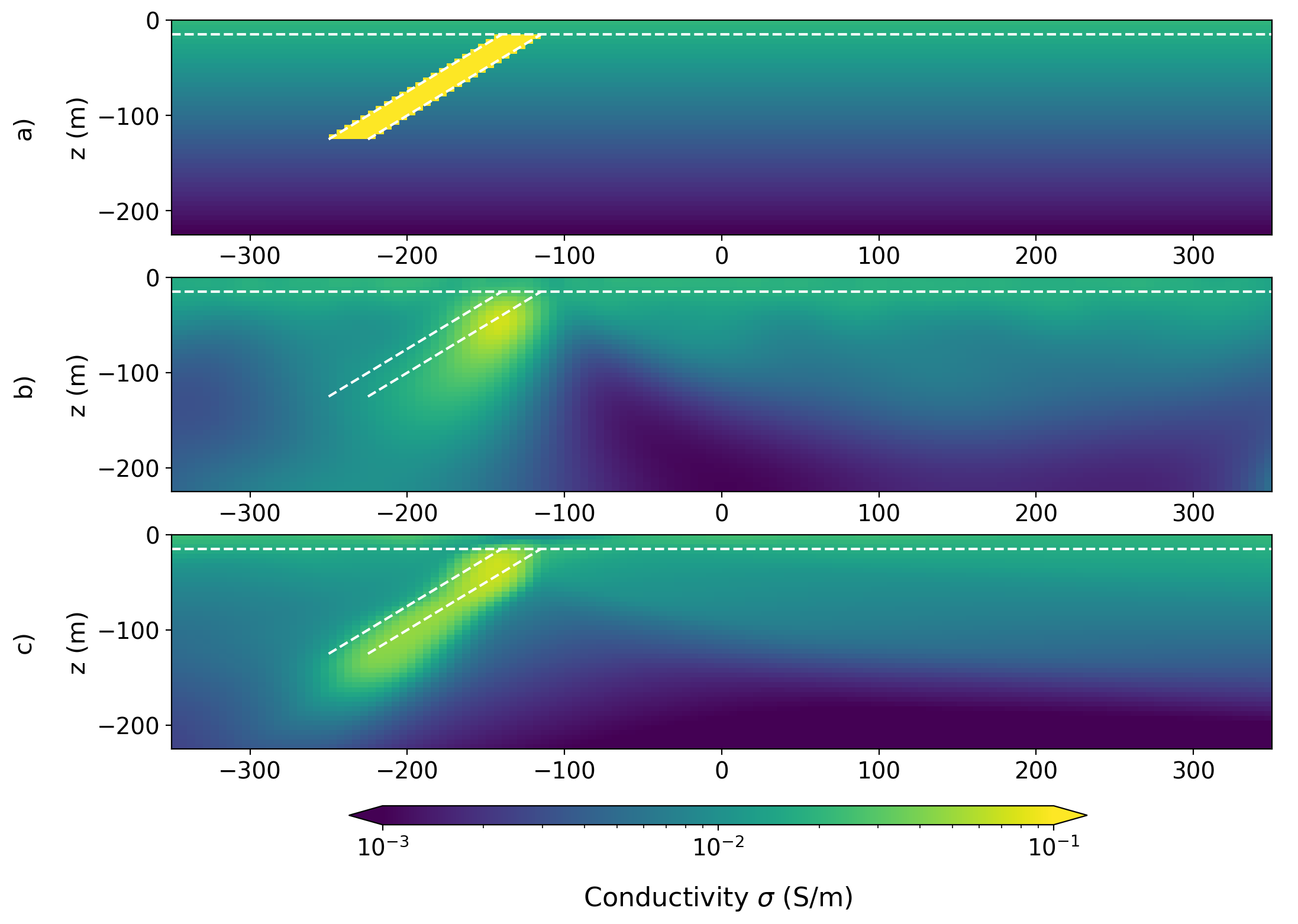}
\caption{(a) is the true model for Case 4. (b) is the recovered model from the conventional method for Case 4 (with L2 smoothness). (c) is the recovered model from the NFs-Inv method for Case 4 (without any explicit regularization).}
\label{NF_Case_4_new}
\end{figure}

To further test the proposed method in a more complicated scenario, we conduct an inversion with the proposed method in a synthetic case with a heterogeneous background [Fig. \ref{NF_Case_4_new}(a)]. The heterogeneous background has a conductivity value that decreases linearly from 0.02 to 0.001 S/m with depth.

Since we assume that we do not have knowledge about the heterogeneous background, we don't have a reference model for Case 4. For the conventional inversion, the smallness term is not applied due to the lack of the reference model, but the smoothness term with $p_x = p_z = 2$ is applied in the inversion process. The starting model is a uniform half-space with a conductivity value of 0.001 S/m. The recovered model from the conventional method can be found in Fig. \ref{NF_Case_4_new}(b).

The recovered model from the proposed method can be found in Fig. \ref{NF_Case_4_new}(c). For the Adam optimizer, the learning rate is 0.001. Although the boundary of the dike structure cannot be recovered as compactly as in Case 3, the location of the main dike is correctly predicted. The conductivity values for the heterogeneous background are predicted approximately correctly, except for the region near the dike structure. Compared to the result from the conventional method, the inversion result from the proposed method has a better depth extent of the dike structure. 

\section{Implicit Bias from the Neural Fields}
\label{sec:4}
In this section, we will explore the implicit regularization effect observed in the results from the proposed method. In Section \ref{sec:4.1}, we illustrate that by searching in a high-dimensional space and leveraging the implicit bias from the neural network, we are able to overcome unwanted artifacts caused by the sensitivity. In Section \ref{sec:4.2}, we elaborate on the choice of positional encoding in different scenarios.

\subsection{Benefits of Changing the Searching Space}
\label{sec:4.1}
\begin{figure}[h!]
\centering
\includegraphics[width=3.5in]{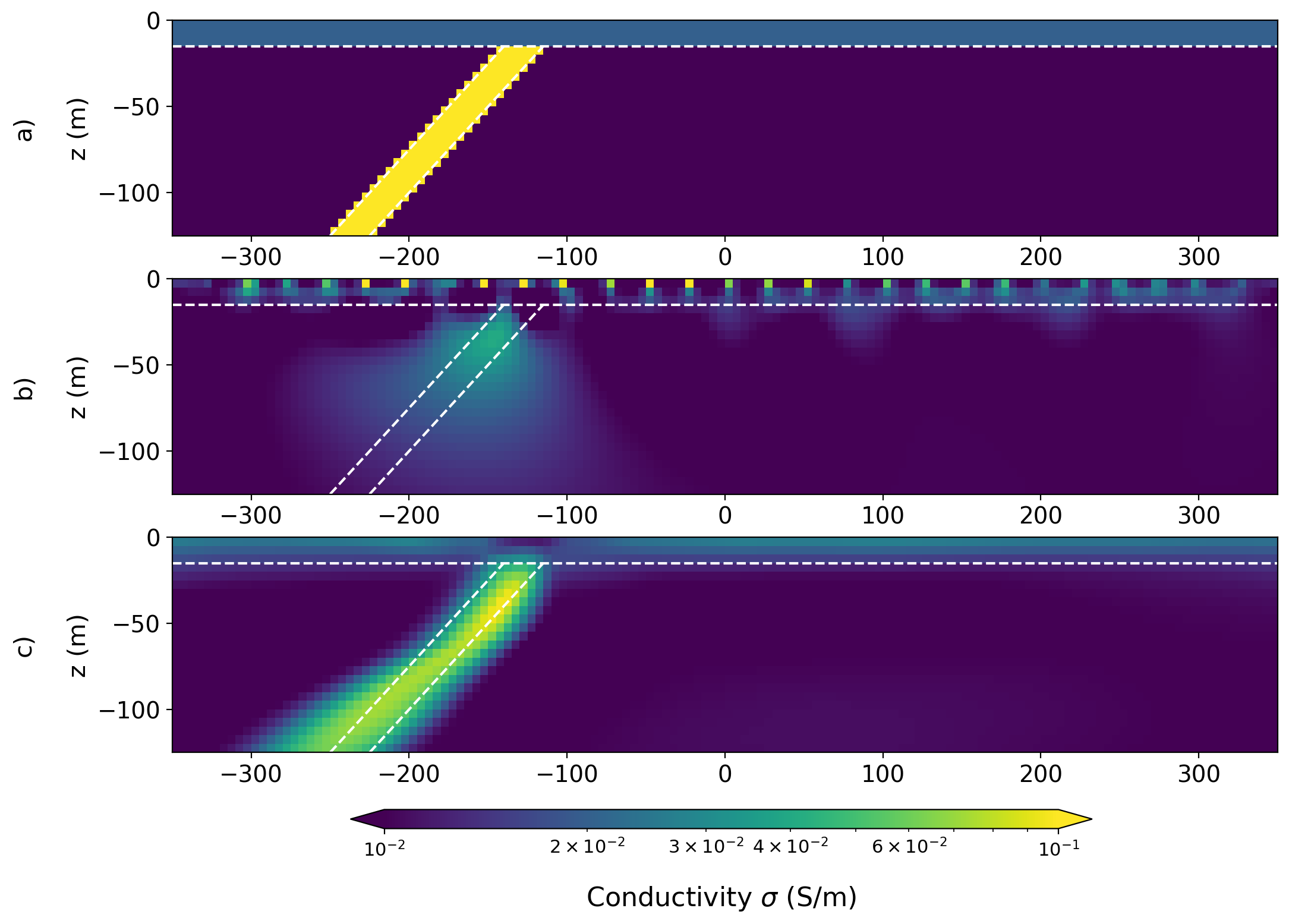}
\caption{(a) is the true model for Case 3. (b) is the recovered model from the conventional method without any explicit regularization for Case 3. (c) is the recovered model from the proposed NFs-Inv method without any explicit regularization for Case 3.}
\label{4_1}
\end{figure}

In Section \ref{sec:3.1}, the results demonstrate that when searching in the space of the NN-weights for a solution, the inversion results are less influenced by sensitivity artifacts. This is especially noticeable when comparing the conventional and NFs-Inv results when no explicit regularization is used [Fig. \ref{4_1}]. Unlike the seismic inversions, the conventional DCR inversion results are not very meaningful if no regularization is included [Fig. \ref{4_1}(b)]. Namely, searching for the solution in the original mesh space is largely affected by the sensitivity, so the results have unrealistic artifacts near the electrodes [Fig. \ref{4_1}(b)] and therefore the conventional approach cannot correctly predict the structures at depth, such as the dike in Case 3. However, even if we do not apply any regularization to the proposed method (i.e. $\beta=0$), we are able to recover the main features, for example, the dike and layered structures in Case 3 [\ref{4_1}(c)]. Considering that the only regularization is from the implicit bias of the proposed method, it is fair to conclude that the neural fields methods have a useful implicit regularization effect on the geophysical inverse problem. The challenge of overcoming artifacts due to the sensitivities of a given survey is common across many geophysical experiments (e.g., it is common to see structures cluster around ground electromagnetic stations). This also indicates the potential application of the NFs-Inv method to other geophysical data types.

\subsection{Apply Positional Encoding to Recover the Higher-frequency Variation}
\label{sec:4.2}

To further test the usage of the positional encoding ($\gamma$), we conduct an ablation study on Cases 2 $\&$ 4 where we apply the proposed method with/without positional encoding. We observe that the choice of positional encoding is problem-dependent. For the results presented in this subsection, we use a Gaussian positional encoding for Case 2 (seismic tomography) and a linear positional encoding for Case 4 (DCR). 

\begin{figure}[h!]
\centering
\includegraphics[width=3.5in]{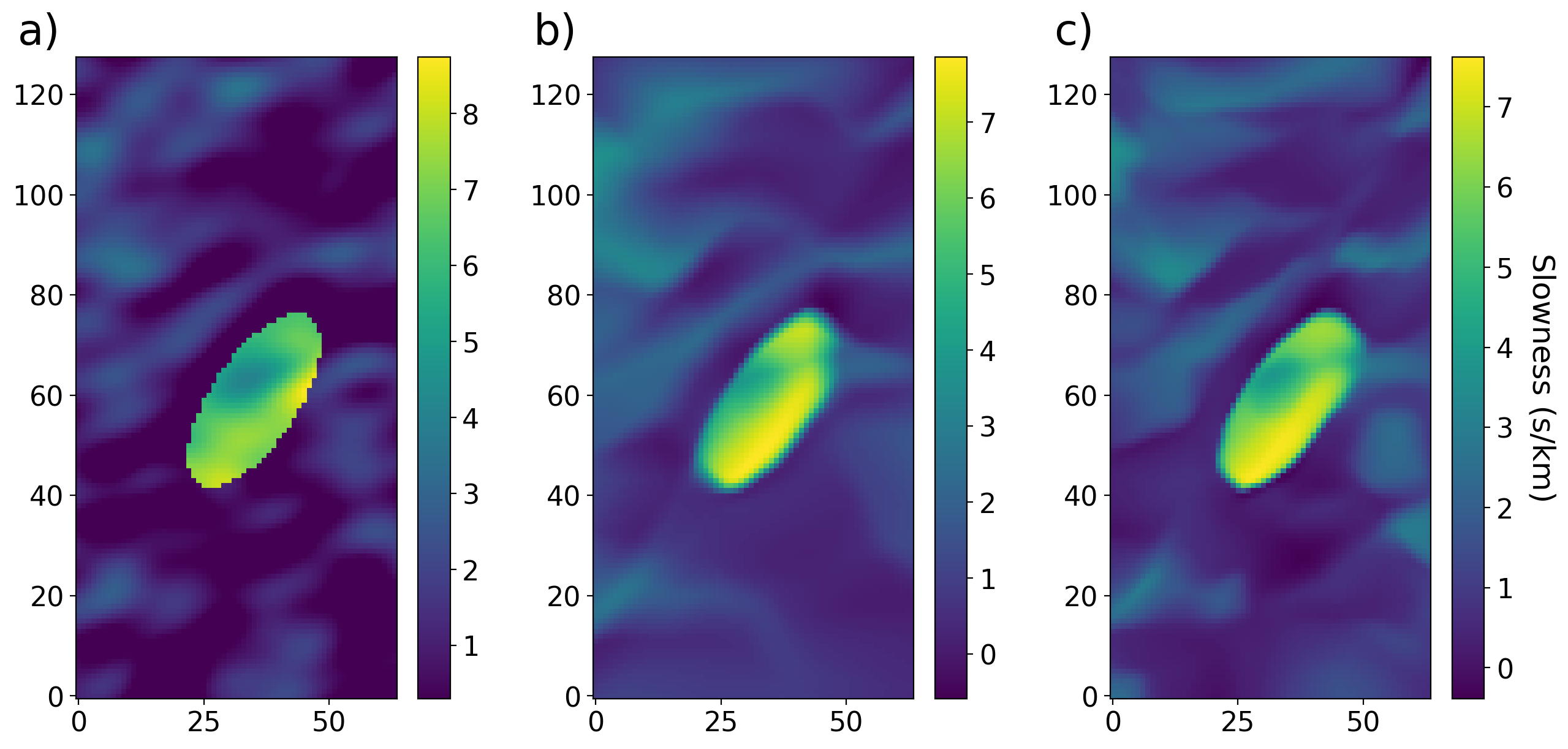}
\caption{(a) is the true model for Case 2. (b) is the proposed NFs-Inv inversion result with an identity $\gamma$. (c) is the proposed NFs-Inv inversion result with a Gaussian positional encoding where $B \in R^{128 \times 2}$ in eqn. \ref{Gaussian encoding}.}
\label{NF_FFT_Case_2}
\end{figure}

\begin{figure}[h!]
\centering
\includegraphics[width=3.5in]{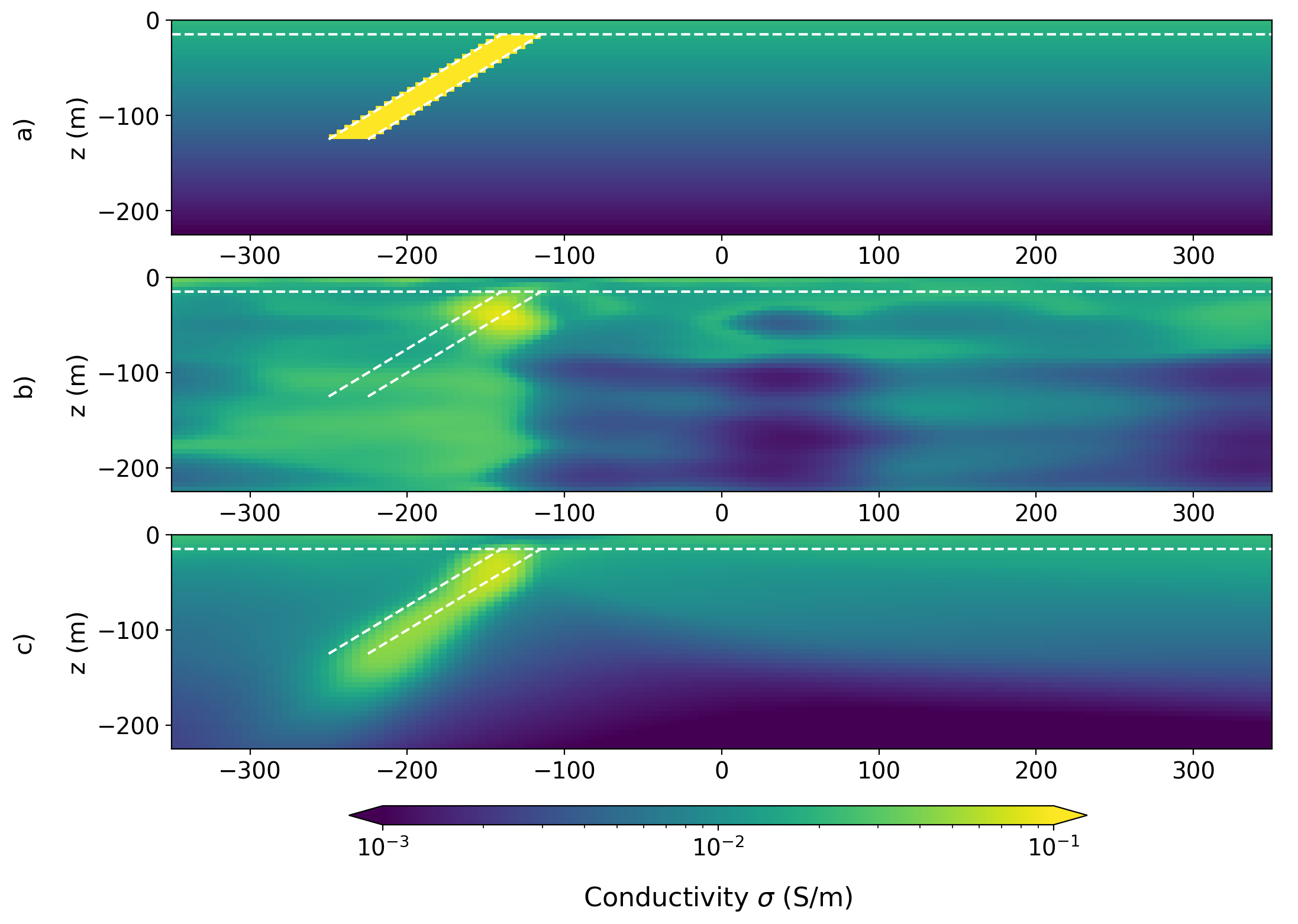}
\caption{((a) is the true model for Case 4. (b) is the proposed NFs-Inv result with a linear positional encoding where m = 8 in eqn. \ref{linear encoding}. (c) is the proposed NFs-Inv result with an identity $\gamma$. }
\label{NF_FFT_Case_4}
\end{figure}

From Fig. \ref{NF_FFT_Case_2}, we observe that without a positional encoding, the recovered model is smoother and details in the background are lost [Fig. \ref{NF_FFT_Case_2}(b)]. By using a positional encoding [Fig. \ref{NF_FFT_Case_2}(c)], we recover a sharper edge to the boundary of the main target and find details in the background slowness model. In contrast, adding positional encoding will give unwanted disturbances to the prediction of the main structures in Case 4 [Fig. \ref{NF_FFT_Case_4}]. Notice that the inversion results in Fig. \ref{NF_FFT_Case_4} fit the observations at the same level, but the recovered models differ a lot. This evidence shows that the $\gamma$ function also introduces an implicit bias to the geophysical inverse problems.

As mentioned before, mapping the inputs to a higher-dimensional space using high-frequency functions prior to passing them to the network enables better fitting of high-frequency variation data. That is why we observe a better recovery of the heterogeneous object in the seismic tomographic example. The fact that we observe the deteriorated recovery of the model in Case 4 illustrates that the choice of positional encoding should align with the ability of the geophysical survey to resolve the subsurface model. Namely, the positional encoding introduces an additional bias to the geophysical inverse problems, which may deteriorate the results in some cases. The resolution of a survey depends upon the underlying physics of the problem, survey geometry, geologic complexity, etc. In the case of cross-hole seismic tomography inversion, the sensors are located on both sides of the survey domain, so there are enough data to constrain the possible solutions. The inversion results with or without the positional encoding can recover similar contours of the subsurface model. Therefore, adding a positional encoding can be helpful for recovering the high-frequency component in the subsurface model in the seismic case. In contrast, the DCR survey for Case 3 $\&$ 4 only has sensors on the ground, and the survey domain is larger than Case 1 $\&$ 2. As a result, Case 3 $\&$ 4 is much more ill-posed than Cases 1 $\&$ 2. Namely, the solution is less constrained, and there are more geologically plausible solutions that can fit the observations to the same degree. In the DCR examples, the additional bias provided by the positional encoding may cause the inversion result to have unrealistically high-frequency signals. As a result, whether we choose to employ a positional encoding should depend on the type of survey method. In terms of selecting the optimal positional encoding, we found that the Gaussian positional encoding with a small standard deviation for generating the matrix $B$ (eqn. (7)) can produce similar results as the basic positional encoding. The selection of the hyperparameters of the positional encoding depends on the complexity of the subsurface model. A more practical guide for choosing positional encodings is an area for future research and will be informed by conducting more trials on different subsurface models and survey methods. 

\section{Analysis using the singular value decomposition analysis of the Jacobian of the Neural Network}
\label{sec:5}
In this section, we perform an empirical analysis to examine the implicit bias that is utilized in the NFs-Inv approach. We hypothesize that this bias is similar to the implicit bias that is utilized in the supervised learning models and which helps generalizability.

While deep learning has demonstrated remarkable success across a growing range of tasks, a comprehensive understanding of the underlying principles driving these achievements remains limited. Many empirical-based methods (``scientific method") have been developed for studying the underlying mechanism \cite{Physics_of_LLMS}, \cite{kaplan2020scalinglawsneurallanguage}, \cite{Zahra}. For example, Kadkhodaie \textit{et al.} \cite{Zahra} find that deep denoisers have inductive biases toward learning a geometry adaptive harmonic basis, and it could potentially lead to solving the generalization puzzle in many large ML models. The authors acknowledge that there must be priors from NN structures and the optimization methods, and the authors refer to it as inductive bias. For clarification, we use implicit bias instead of inductive bias in our context to avoid potential ambiguity since some papers refer to inductive bias in CNN specifically as translational invariance \cite{kayhan2020translationinvariancecnnsconvolutional}. Kadkhodaie \textit{et al.} \cite{Zahra} state that since a DNN can be considered as a locally linear function, conducting an eigendecomposition analysis (SVD analysis) of the Jacobian of the neural network concerning the input can reveal what has been learned by the generative models. Namely, since these left-singular vectors form the basis of the output, the features of the basis can indicate what has been imposed by the implicit bias during the training process. They find that the learned bases have oscillating harmonic structures along contours and in homogeneous regions. They name these bases as geometry-adaptive harmonic bases (GAHBs). 

In our work, we also find that the Jacobian has a geometry-adaptive harmonic basis after inversion. Note that Kadkhodaie \textit{et al.} \cite{Zahra} conducted an SVD analysis of the Jacobian of the output with respect to the input. In their supervised learning setting, the inputs of the model are different for different input images. However, in our case, since the inputs are always the fixed values of the coordinates of each cell, we conduct an SVD analysis of the Jacobian of the output with respect to the weights of the NN $f$. Since the choice of the positional encoding may introduce an additional bias (See Section \ref{sec:4.2}), we do not employ positional encoding for the models presented in this subsection (i.e., $\gamma$ is an identity function) to avoid potential ambiguity. The SVD of the Jacobian is

\begin{equation}
\label{SVD of J}
J = \frac{\partial m}{\partial w} = U \Lambda V^T
\end{equation}
The left-singular vectors (the columns of $U$) have the dimensionality of the geophysical model, and the right-singular vectors (the columns of $V$) have the dimensionality of the number of weights in the neural network. $\Lambda$ is a diagonal matrix whose elements are the singular values $\lambda_i$. We focus our analysis on the left-singular vectors, as these can be thought of as the learned basis for our geophysical model. 

\begin{figure}[h!]
\centering
\includegraphics[width=3.in]{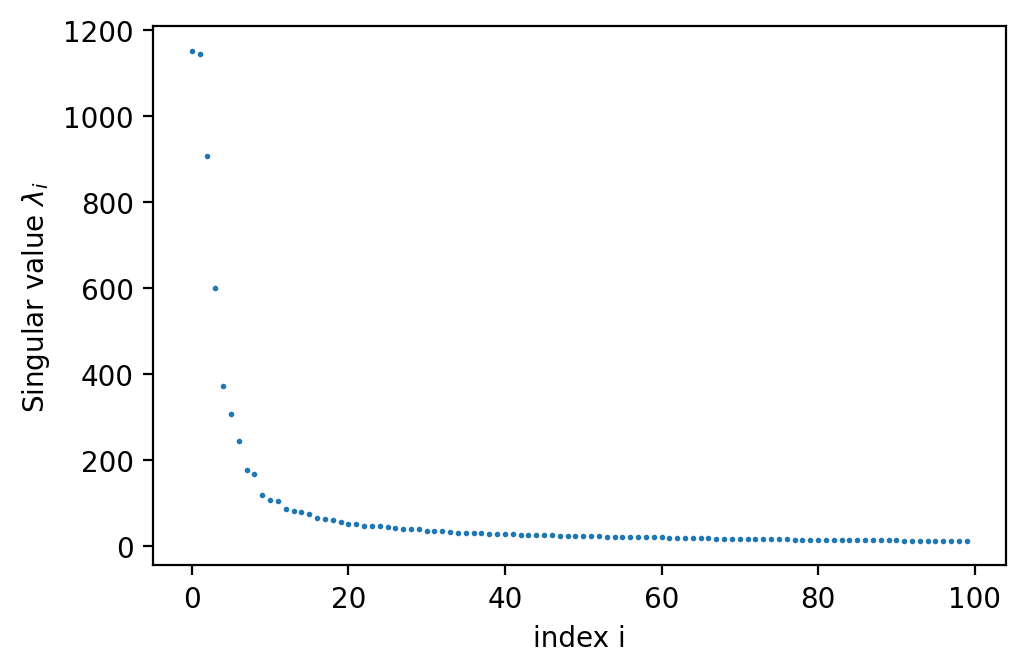}
\includegraphics[width=3.5in]{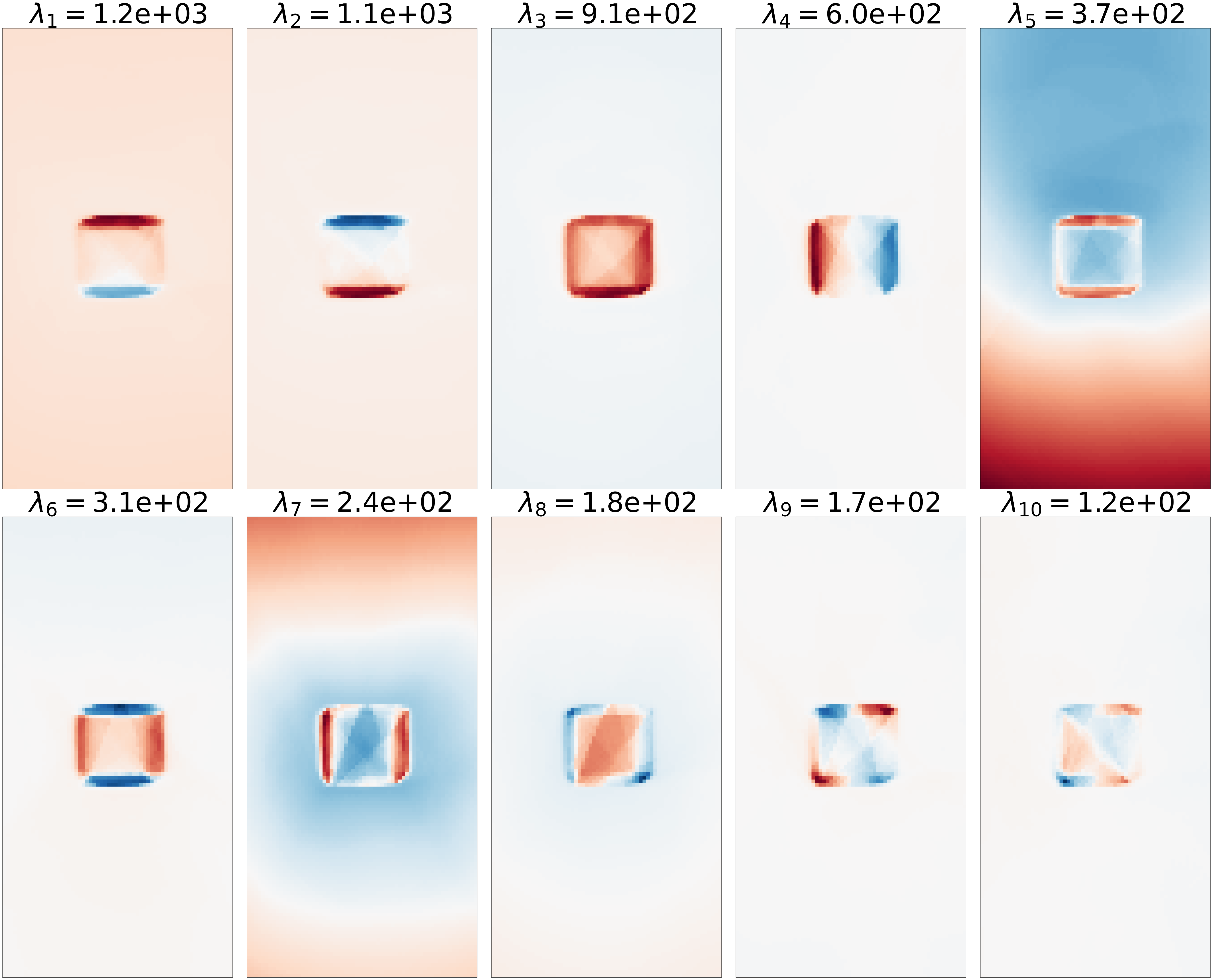}
\caption{First 100 singular values and first ten left-singular vectors for the Jacobian in Case 1.}
\label{Case_1_eigen}
\end{figure}

\begin{figure}[h!]
\centering
\includegraphics[width=3.in]{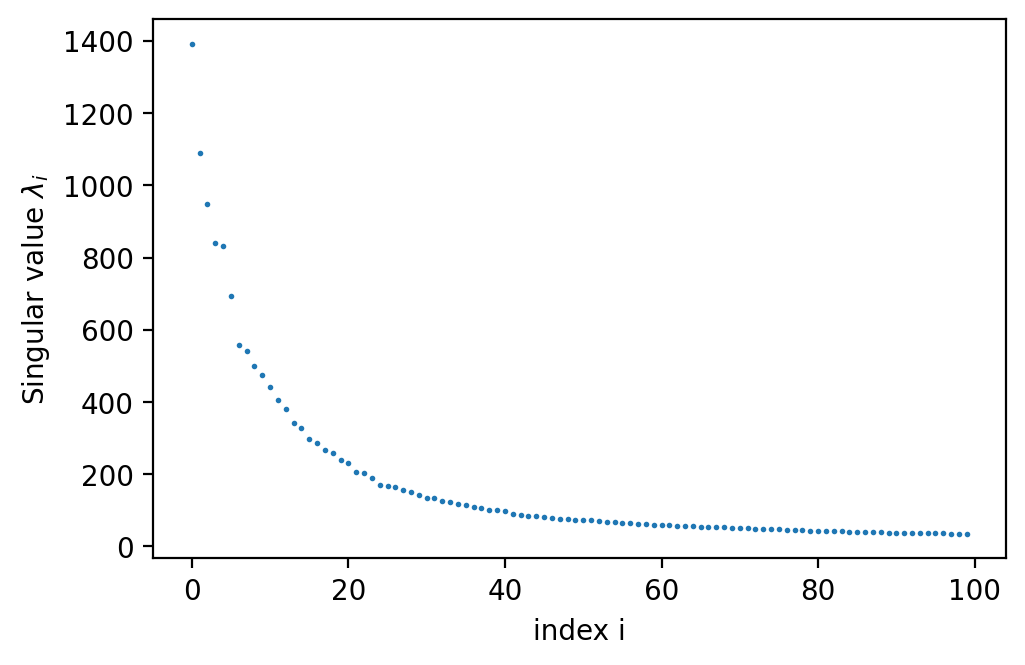}
\includegraphics[width=3.5in]{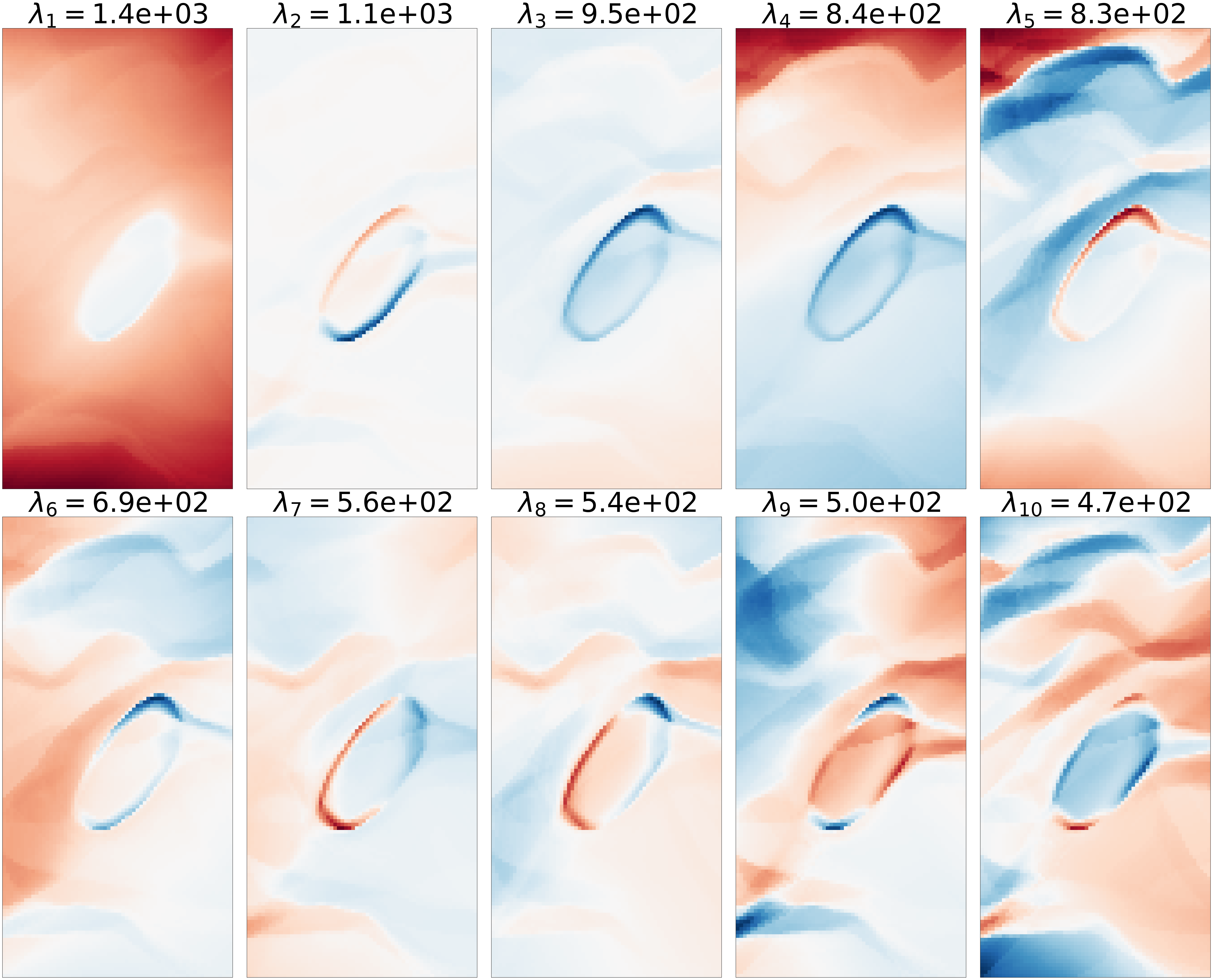}
\caption{First 100 singular values and first ten left-singular vectors for the Jacobian in Case 2.}
\label{Case_2_eigen}
\end{figure}

Fig. \ref{Case_1_eigen} shows the first ten left-singular vectors for Case 1. As observed in Kadkhodaie \textit{et al.} \cite{Zahra}, the first few left-singular vectors strongly highlight the edges of the main target, while the background is relatively homogeneous (or smooth in cases where there is structure). Fig. \ref{Case_2_eigen} shows the first ten left-singular vectors for Case 2. We can observe that the left-singular vectors capture the main geometric structures related to the elliptical target. The first three left-singular vectors highlight the main elliptical target and are relatively homogeneous in the background. Starting from the fourth left-singular vector, we then see more structures associated with details and higher frequency variation in the background. 

\begin{figure}[h!]
\centering
\includegraphics[width=3.in]{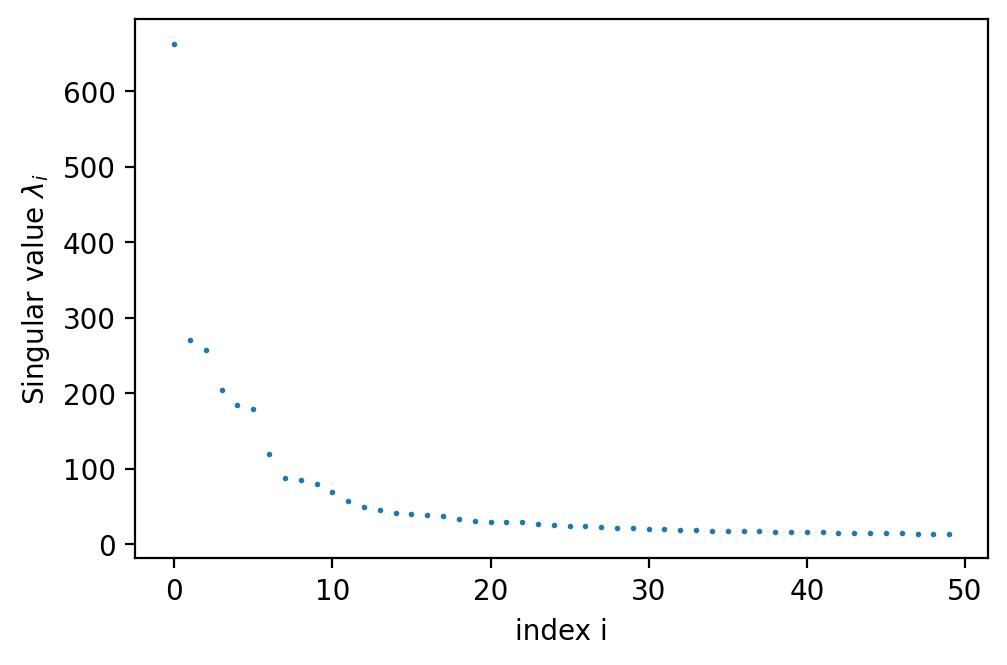}
\includegraphics[width=3.5in]{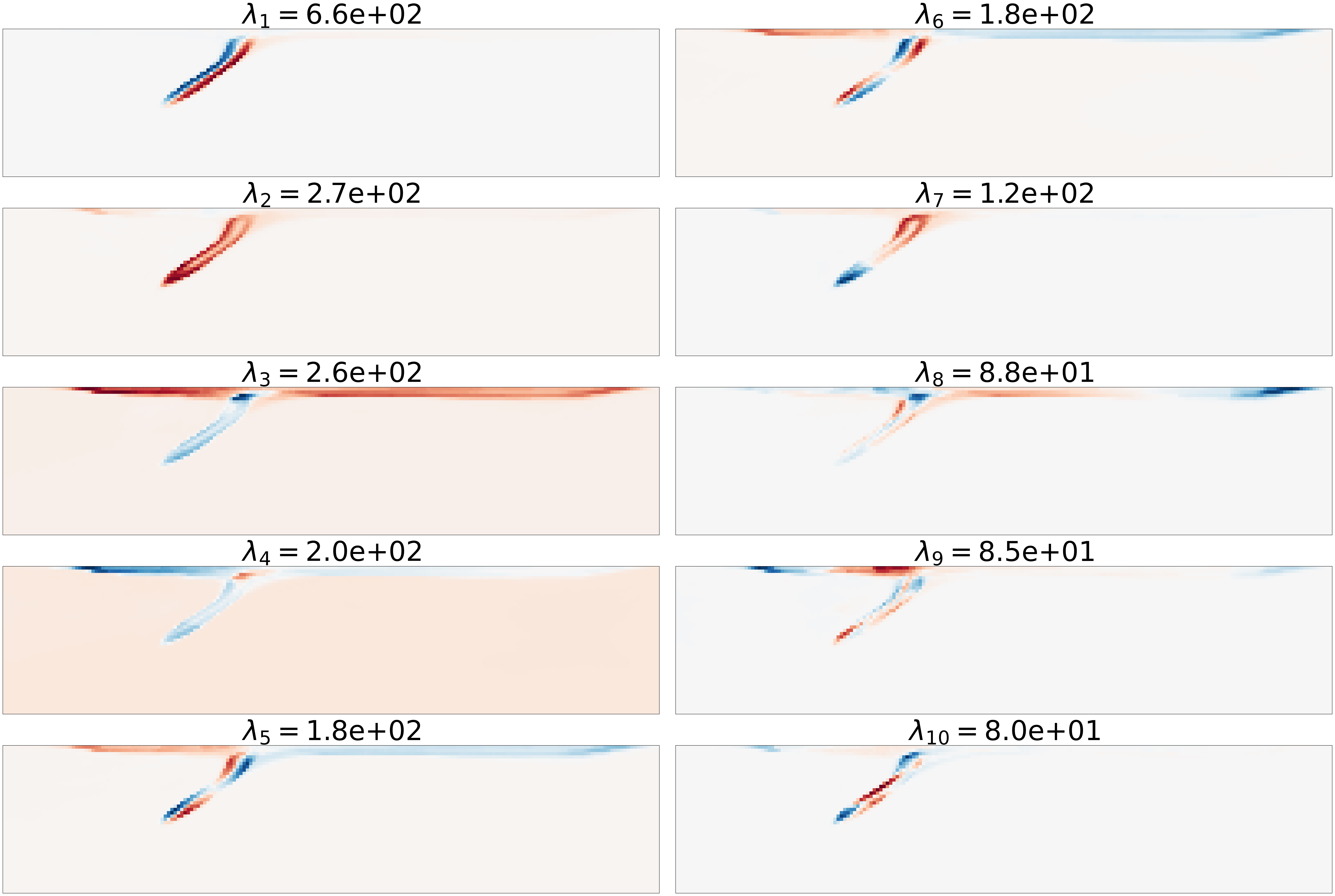}
\caption{First 50 singular values and first ten left-singular vectors for the Jacobian in Case 3.}
\label{Case_3_eigen}
\end{figure}

\begin{figure}[h!]
\centering
\includegraphics[width=3.in]{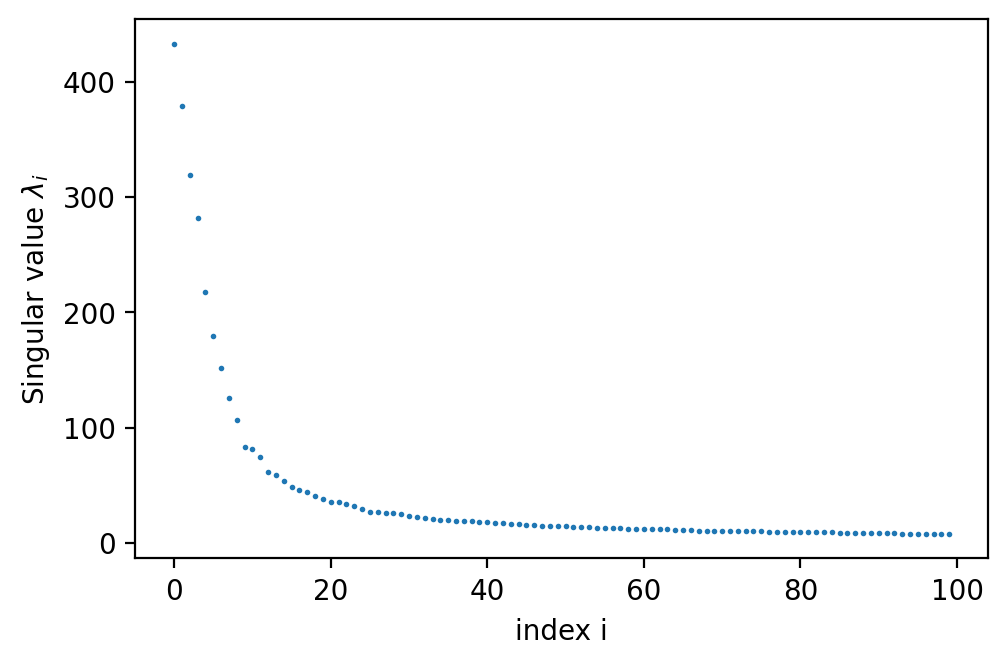}
\includegraphics[width=3.5in]{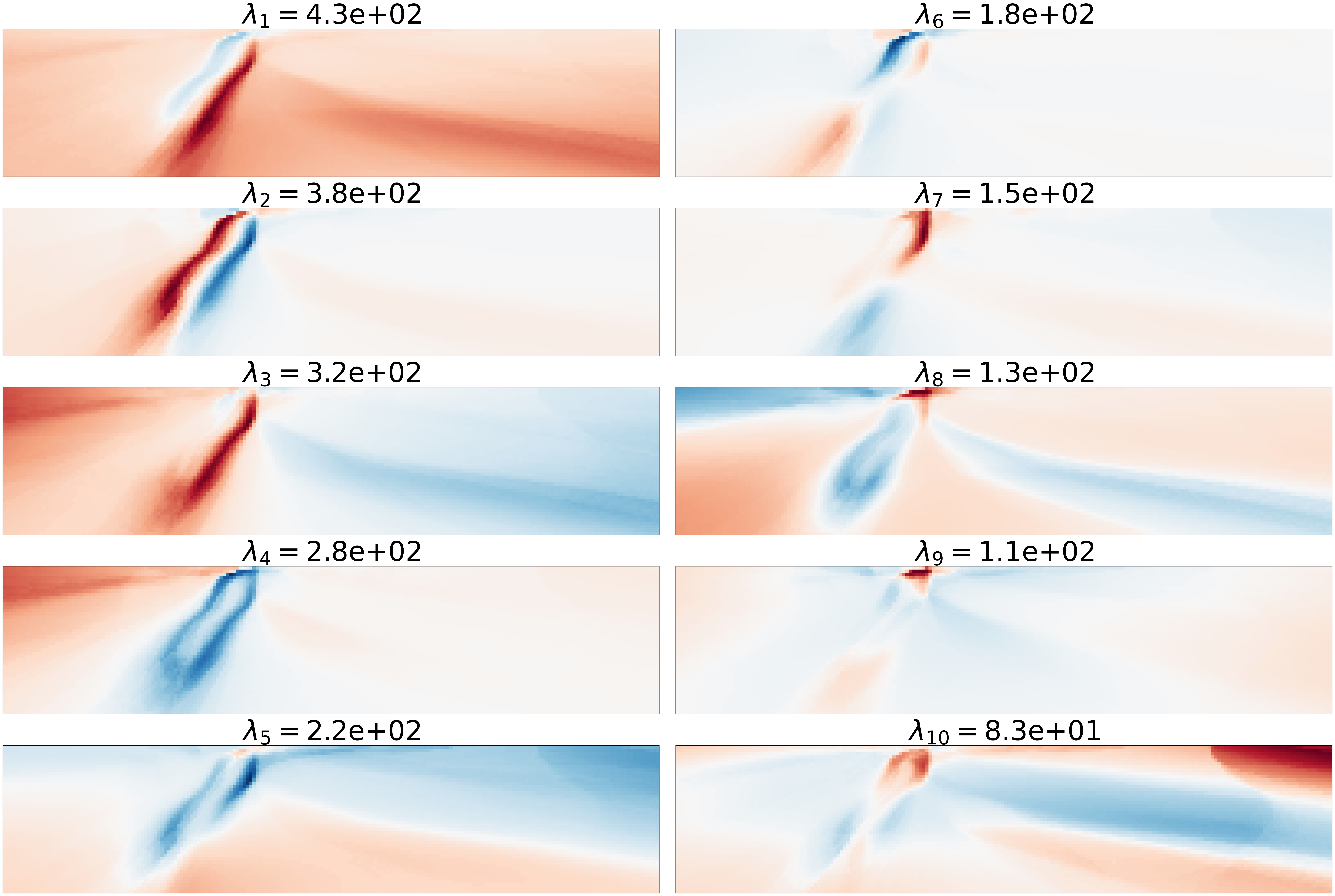}
\caption{First 100 singular values and first ten left-singular vectors for the Jacobian in Case 4.}
\label{Case_4_eigen}
\end{figure}

Fig. \ref{Case_3_eigen} and Fig. \ref{Case_4_eigen} show the first ten left-singular vectors for the DCR inversion cases. Similarly, we observe harmonic features along the contour of the main dike target in the DCR case. The first two left-singular vectors highlight the dike structure. Starting from the third left-singular vector, we also see the influence of the near-surface layer. 

From the plots of the eigenvalues for all 4 examples, we find that the eigenvalues are sparse. Namely, the first few eigenvalues are significantly larger than the remaining eigenvalues. In Case 1 and Case 3, which are both relatively simple, there is a reduction of nearly a factor of 10 between the first and tenth singular values. For Case 2, which is more complex, there is a reduction of a factor of around 3 between the first and tenth singular values. This observation agrees with the finding in the analysis of Jacobian in the diffusion models \cite{Zahra}, where they also find this shrinkage of the eigenvalues.

These examples are evidence that there exists an implicit bias, which has been illustrated in supervised learning models and may also be utilized in test-time learning models as well. Moreover, this analysis can partly explain the reason why we can observe a better recovery of the subsurface physical properties when we re-parameterize the subsurface model using the weights of a DNN. By forming these adaptive bases, we have the chance to overcome the problems of predicting unwanted artifacts due to sensitivity, such as near-electrode artifacts or the concentration along the ray path.

\section{Discussion and Future Work}

Further work, such as the applications to the field data and other theoretical analyses of the implicit bias, is still in progress. It has been shown that using the adaptive grids can improve the performance of the neural fields methods \cite{martel2021acornadaptivecoordinatenetworks}, \cite{10.1145/3503250}, so testing the adaptive grid scheme for geophysical inversions is a possible future research direction.

The run-time needed for convergence of the NFs-Inv algorithm heavily depends on the time required to run the forward simulation. For the seismic tomography examples, since the problem is linear can run on a GPU, the computational costs for both the conventional and proposed methods are small and similar. For the DCR inversion, the runtime between the conventional and the proposed method differs significantly. The CPU time is averaged over 4 trials running on a 48-core CPU server with Intel(R) Xeon(R) CPU E5-2670 v3 $@$ 2.30GHz. The computational time for the conventional method using sparse-norm inversion with 44 epochs is 466 seconds. For our proposed approach, we ran for 2000 epochs; however, in practice, there was minimal difference between the solution observed at 1000 epochs and the final solution. The computational time for the proposed method with 1000 epochs is 9708 seconds. The time per epoch is similar for both approaches, as 98.5$\%$ of the time is spent on the forward mapping and calculation of the Jacobian matrix using SimPEG. Future work on improving the computational cost includes designing a suitable optimization to decrease the number of epochs needed and improving the efficiency of the forward simulation. 

The results shown in Section \ref{sec:3} are state-of-the-art, which we established by testing the performance of different positional encoding functions. We found that in the cases where the subsurface model is relatively simple, such as Cases 1 $\&$ 3, adding a positional encoding function where $h$ is large will deteriorate the inversion results. However, the results of Case 2 show that when the model is more complex and we use a geophysical method with sufficient sensitivity, adding positional encoding can help us to recover sharper boundaries and finer details in the model. There is still some work left on exploring the effect of the choice of the $\gamma$ function on the proposed method.  

We have tested employing a sine function \cite{SIREN} as the activation function in Cases 2 $\&$ 4, and the results are not ideal since it decreased the sharpness of the boundaries of the main target. Kadkhodaie \textit{et al.} \cite{Zahra} found that ReLUs are key for observing geometric adaptivity in the SVD analysis on diffusion models. We hypothesize that the Leaky Rectified Linear Unit activation layer (LeakyReLU) is potentially a critical component of the implicit bias we observed. More tests on the performance of the adaptive activation function should be done in the future.
 
\section{Conclusion}

In conclusion, the application of neural fields in geophysical inverse problems, as explored in this work, demonstrates significant advantages over conventional inversion methods. By framing the inverse problem in terms of Neural Network (NN) weights, we leverage the high-dimensional search space of the weights. Using a coordinate-based network allows us to take advantage of the built-in smoothing effects that come from parameterizing the inverse problem in a continuous setting. The results we showed, using seismic tomography and direct current resistivity, highlight the ability of neural fields to reduce unwanted artifacts and improve the accuracy of subsurface models, particularly in capturing the boundaries and dip angles of key targets. This test-time learning approach, driven by the implicit bias of neural fields, proves to be a promising direction for enhancing inversion results in geophysical problems.

Moreover, using an empirical approach, we provided evidence that there exists an implicit bias, which has been illustrated in supervised learning, that can also be utilized in test-time learning models. By forming an adaptive basis during the inversion, we have the chance to overcome the problems of predicting unwanted artifacts due to sensitivity, such as near-electrode artifacts (DCR) or the concentration along the ray path (seismic). This analysis can partly explain the reason why we can observe a better recovery of the subsurface physical properties when we re-parameterize the subsurface model using the weights of a DNN. By showing that the implicit bias brought by the DNNs can benefit geophysical inversions, we also give insights into other machine learning methods in geophysical problems.

\section*{Data Availability}
The code used is available at \\https://doi.org/10.5281/zenodo.15034077.

\section*{Acknowledgments}
This research was supported by funding from the NSERC Discovery Grants Program and the Mitacs Accelerate Program. We would like to thank Zahra Kadkhodaie for her elaboration of her works. We also want to thank UBC GIF group members for their invaluable guidance and advice on the usage of SimPEG. We are grateful for the careful review and constructive feedback provided by the reviewers.

\bibliography{IEEE_NFs_Inv}{}

\begin{thebibliography}{10}
\providecommand{\url}[1]{#1}
\csname url@samestyle\endcsname
\providecommand{\newblock}{\relax}
\providecommand{\bibinfo}[2]{#2}
\providecommand{\BIBentrySTDinterwordspacing}{\spaceskip=0pt\relax}
\providecommand{\BIBentryALTinterwordstretchfactor}{4}
\providecommand{\BIBentryALTinterwordspacing}{\spaceskip=\fontdimen2\font plus
\BIBentryALTinterwordstretchfactor\fontdimen3\font minus \fontdimen4\font\relax}
\providecommand{\BIBforeignlanguage}[2]{{%
\expandafter\ifx\csname l@#1\endcsname\relax
\typeout{** WARNING: IEEEtran.bst: No hyphenation pattern has been}%
\typeout{** loaded for the language `#1'. Using the pattern for}%
\typeout{** the default language instead.}%
\else
\language=\csname l@#1\endcsname
\fi
#2}}
\providecommand{\BIBdecl}{\relax}
\BIBdecl

\bibitem{menke_1989}
W.~Menke, \emph{Geophysical Data Analysis}.\hskip 1em plus 0.5em minus 0.4em\relax Elsevier, Oct 1989.

\bibitem{ref47}
D.~Oldenburg and Y.~Li, ``\BIBforeignlanguage{en}{Inversion for applied geophysics: A tutorial},'' \emph{\BIBforeignlanguage{en}{Investigations in geophysics}}, vol.~13, pp. 1–85,, 2005.

\bibitem{cowan_airborne_2025}
\BIBentryALTinterwordspacing
D.~C. Cowan, L.~J. Heagy, and D.~W. Oldenburg, ``Airborne natural source electromagnetics for an arbitrary base station,'' \emph{Geophysical Journal International}, vol. 241, no.~3, pp. 1962--1975, Apr. 2025, \_eprint: https://academic.oup.com/gji/article-pdf/241/3/1962/63015253/ggaf059.pdf. [Online]. Available: \url{https://doi.org/10.1093/gji/ggaf059}
\BIBentrySTDinterwordspacing

\bibitem{fournier_inversion_2019}
\BIBentryALTinterwordspacing
D.~Fournier and D.~W. Oldenburg, ``Inversion using spatially variable mixed $l$p norms,'' \emph{Geophysical Journal International}, vol. 218, no.~1, pp. 268--282, Mar. 2019, \_eprint: https://academic.oup.com/gji/article-pdf/218/1/268/28461599/ggz156.pdf. [Online]. Available: \url{https://doi.org/10.1093/gji/ggz156}
\BIBentrySTDinterwordspacing

\bibitem{Soler2025-kh}
S.~Soler, J.~Capriotti, D.~Oldenburg, and L.~Heagy, ``3d geophysical inversions to characterize carbon sequestration potential of ultramafic rocks,'' mar 2025.

\bibitem{https://doi.org/10.1029/2021JB022581}
\BIBentryALTinterwordspacing
J.~Lopez-Alvis, F.~Nguyen, M.~C. Looms, and T.~Hermans, ``Geophysical inversion using a variational autoencoder to model an assembled spatial prior uncertainty,'' \emph{Journal of Geophysical Research: Solid Earth}, vol. 127, no.~3, p. e2021JB022581, 2022, e2021JB022581 2021JB022581. [Online]. Available: \url{https://agupubs.onlinelibrary.wiley.com/doi/abs/10.1029/2021JB022581}
\BIBentrySTDinterwordspacing

\bibitem{alyousuf_threeaxis_2024}
\BIBentryALTinterwordspacing
T.~Alyousuf, Y.~Li, R.~Krahenbuhl, and D.~Grana, ``\BIBforeignlanguage{en}{Three‐axis borehole gravity monitoring for {CO} $_{\textrm{2}}$ storage using machine learning coupled to fluid flow simulator},'' \emph{\BIBforeignlanguage{en}{Geophysical Prospecting}}, vol.~72, no.~2, pp. 767--790, Feb. 2024. [Online]. Available: \url{https://onlinelibrary.wiley.com/doi/10.1111/1365-2478.13413}
\BIBentrySTDinterwordspacing

\bibitem{zhang_velocitygan_2019}
\BIBentryALTinterwordspacing
Z.~Zhang, Y.~Wu, Z.~Zhou, and Y.~Lin, ``\BIBforeignlanguage{en}{{VelocityGAN}: {Subsurface} {Velocity} {Image} {Estimation} {Using} {Conditional} {Adversarial} {Networks}},'' in \emph{\BIBforeignlanguage{en}{2019 {IEEE} {Winter} {Conference} on {Applications} of {Computer} {Vision} ({WACV})}}.\hskip 1em plus 0.5em minus 0.4em\relax Waikoloa Village, HI, USA: IEEE, Jan. 2019, pp. 705--714. [Online]. Available: \url{https://ieeexplore.ieee.org/document/8658391/}
\BIBentrySTDinterwordspacing

\bibitem{8994191}
B.~Liu, Q.~Guo, S.~Li, B.~Liu, Y.~Ren, Y.~Pang, X.~Guo, L.~Liu, and P.~Jiang, ``Deep learning inversion of electrical resistivity data,'' \emph{IEEE Transactions on Geoscience and Remote Sensing}, vol.~58, no.~8, pp. 5715--5728, 2020.

\bibitem{https://doi.org/10.1111/cgf.14505}
\BIBentryALTinterwordspacing
Y.~Xie, T.~Takikawa, S.~Saito, O.~Litany, S.~Yan, N.~Khan, F.~Tombari, J.~Tompkin, V.~sitzmann, and S.~Sridhar, ``Neural fields in visual computing and beyond,'' \emph{Computer Graphics Forum}, vol.~41, no.~2, pp. 641--676, 2022. [Online]. Available: \url{https://onlinelibrary.wiley.com/doi/abs/10.1111/cgf.14505}
\BIBentrySTDinterwordspacing

\bibitem{10.1145/3503250}
\BIBentryALTinterwordspacing
B.~Mildenhall, P.~P. Srinivasan, M.~Tancik, J.~T. Barron, R.~Ramamoorthi, and R.~Ng, ``Nerf: representing scenes as neural radiance fields for view synthesis,'' \emph{Commun. ACM}, vol.~65, no.~1, p. 99–106, dec 2021. [Online]. Available: \url{https://doi.org/10.1145/3503250}
\BIBentrySTDinterwordspacing

\bibitem{9709943}
A.~W. Reed, H.~Kim, R.~Anirudh, K.~A. Mohan, K.~Champley, J.~Kang, and S.~Jayasuriya, ``Dynamic ct reconstruction from limited views with implicit neural representations and parametric motion fields,'' in \emph{2021 IEEE/CVF International Conference on Computer Vision (ICCV)}, 2021, pp. 2238--2248.

\bibitem{9606601}
Y.~Sun, J.~Liu, M.~Xie, B.~Wohlberg, and U.~S. Kamilov, ``Coil: Coordinate-based internal learning for tomographic imaging,'' \emph{IEEE Transactions on Computational Imaging}, vol.~7, pp. 1400--1412, 2021.

\bibitem{Cachia_2023}
\BIBentryALTinterwordspacing
M.~Cachia, V.~Stergiopoulou, L.~Calatroni, S.~Schaub, and L.~Blanc-F{\'e}raud, ``Fluorescence image deconvolution microscopy via generative adversarial learning (fluogan),'' \emph{Inverse Problems}, vol.~39, no.~5, p. 054006, apr 2023. [Online]. Available: \url{https://dx.doi.org/10.1088/1361-6420/acc889}
\BIBentrySTDinterwordspacing

\bibitem{yu_plenoxels_2021}
\BIBentryALTinterwordspacing
A.~Yu, S.~Fridovich-Keil, M.~Tancik, Q.~Chen, B.~Recht, and A.~Kanazawa, ``\BIBforeignlanguage{en}{Plenoxels: {Radiance} {Fields} without {Neural} {Networks}},'' Dec. 2021, arXiv:2112.05131 [cs]. [Online]. Available: \url{http://arxiv.org/abs/2112.05131}
\BIBentrySTDinterwordspacing

\bibitem{martinbrualla2021nerfwildneuralradiance}
\BIBentryALTinterwordspacing
R.~Martin-Brualla, N.~Radwan, M.~S.~M. Sajjadi, J.~T. Barron, A.~Dosovitskiy, and D.~Duckworth, ``Nerf in the wild: Neural radiance fields for unconstrained photo collections,'' 2021. [Online]. Available: \url{https://arxiv.org/abs/2008.02268}
\BIBentrySTDinterwordspacing

\bibitem{tancik2020fourierfeaturesletnetworks}
\BIBentryALTinterwordspacing
M.~Tancik, P.~P. Srinivasan, B.~Mildenhall, S.~Fridovich-Keil, N.~Raghavan, U.~Singhal, R.~Ramamoorthi, J.~T. Barron, and R.~Ng, ``Fourier features let networks learn high frequency functions in low dimensional domains,'' 2020. [Online]. Available: \url{https://arxiv.org/abs/2006.10739}
\BIBentrySTDinterwordspacing

\bibitem{DBLP}
\BIBentryALTinterwordspacing
J.~Zheng, S.~Ramasinghe, and S.~Lucey, ``Rethinking positional encoding,'' \emph{CoRR}, vol. abs/2107.02561, 2021. [Online]. Available: \url{https://arxiv.org/abs/2107.02561}
\BIBentrySTDinterwordspacing

\bibitem{sitzmann2020implicitneuralrepresentationsperiodic}
\BIBentryALTinterwordspacing
V.~Sitzmann, J.~N.~P. Martel, A.~W. Bergman, D.~B. Lindell, and G.~Wetzstein, ``Implicit neural representations with periodic activation functions,'' 2020. [Online]. Available: \url{https://arxiv.org/abs/2006.09661}
\BIBentrySTDinterwordspacing

\bibitem{fathony2021multiplicative}
\BIBentryALTinterwordspacing
R.~Fathony, A.~K. Sahu, D.~Willmott, and J.~Z. Kolter, ``Multiplicative filter networks,'' in \emph{International Conference on Learning Representations}, 2021. [Online]. Available: \url{https://openreview.net/forum?id=OmtmcPkkhT}
\BIBentrySTDinterwordspacing

\bibitem{arratia2024enhancingdynamicctimage}
\BIBentryALTinterwordspacing
P.~Arratia, M.~Ehrhardt, and L.~Kreusser, ``Enhancing dynamic ct image reconstruction with neural fields through explicit motion regularizers,'' 2024. [Online]. Available: \url{https://arxiv.org/abs/2406.01299}
\BIBentrySTDinterwordspacing

\bibitem{molaei2023implicitneuralrepresentationmedical}
\BIBentryALTinterwordspacing
A.~Molaei, A.~Aminimehr, A.~Tavakoli, A.~Kazerouni, B.~Azad, R.~Azad, and D.~Merhof, ``Implicit neural representation in medical imaging: A comparative survey,'' 2023. [Online]. Available: \url{https://arxiv.org/abs/2307.16142}
\BIBentrySTDinterwordspacing

\bibitem{shen_nerp_2024}
\BIBentryALTinterwordspacing
L.~Shen, J.~Pauly, and L.~Xing, ``\BIBforeignlanguage{en}{{NeRP}: {Implicit} {Neural} {Representation} {Learning} {With} {Prior} {Embedding} for {Sparsely} {Sampled} {Image} {Reconstruction}},'' \emph{\BIBforeignlanguage{en}{IEEE Trans. Neural Netw. Learning Syst.}}, vol.~35, no.~1, pp. 770--782, Jan. 2024. [Online]. Available: \url{https://ieeexplore.ieee.org/document/9788018/}
\BIBentrySTDinterwordspacing

\bibitem{liu2022recoverycontinuous3drefractive}
\BIBentryALTinterwordspacing
R.~Liu, Y.~Sun, J.~Zhu, L.~Tian, and U.~Kamilov, ``Recovery of continuous 3d refractive index maps from discrete intensity-only measurements using neural fields,'' 2022. [Online]. Available: \url{https://arxiv.org/abs/2112.00002}
\BIBentrySTDinterwordspacing

\bibitem{https://doi.org/10.1029/2021JB023120}
\BIBentryALTinterwordspacing
M.~Rasht-Behesht, C.~Huber, K.~Shukla, and G.~E. Karniadakis, ``Physics-informed neural networks (pinns) for wave propagation and full waveform inversions,'' \emph{Journal of Geophysical Research: Solid Earth}, vol. 127, no.~5, p. e2021JB023120, 2022, e2021JB023120 2021JB023120. [Online]. Available: \url{https://agupubs.onlinelibrary.wiley.com/doi/abs/10.1029/2021JB023120}
\BIBentrySTDinterwordspacing

\bibitem{IFWI}
\BIBentryALTinterwordspacing
J.~Sun, K.~Innanen, T.~Zhang, and D.~Trad, ``Implicit seismic full waveform inversion with deep neural representation,'' \emph{Journal of Geophysical Research: Solid Earth}, vol. 128, no.~3, p. e2022JB025964, 2023, e2022JB025964 2022JB025964. [Online]. Available: \url{https://agupubs.onlinelibrary.wiley.com/doi/abs/10.1029/2022JB025964}
\BIBentrySTDinterwordspacing

\bibitem{Mingliang}
\BIBentryALTinterwordspacing
M.~Liu, D.~Vashisth, D.~Grana, and T.~Mukerji, ``Joint inversion of geophysical data for geologic carbon sequestration monitoring: A differentiable physics-informed neural network model,'' \emph{Journal of Geophysical Research: Solid Earth}, vol. 128, no.~3, p. e2022JB025372, 2023, e2022JB025372 2022JB025372. [Online]. Available: \url{https://agupubs.onlinelibrary.wiley.com/doi/abs/10.1029/2022JB025372}
\BIBentrySTDinterwordspacing

\bibitem{https://doi.org/10.1029/2025JH000621}
\BIBentryALTinterwordspacing
Y.~Wu and J.~Ma, ``How does neural network reparametrization improve geophysical inversion?'' \emph{Journal of Geophysical Research: Machine Learning and Computation}, vol.~2, no.~2, p. e2025JH000621, 2025, e2025JH000621 2025JH000621. [Online]. Available: \url{https://agupubs.onlinelibrary.wiley.com/doi/abs/10.1029/2025JH000621}
\BIBentrySTDinterwordspacing

\bibitem{https://doi.org/10.1029/2024JH000542}
\BIBentryALTinterwordspacing
F.~Liu, H.~Li, G.~Zou, and J.~Li, ``Automatic differentiation-based full waveform inversion with flexible workflows,'' \emph{Journal of Geophysical Research: Machine Learning and Computation}, vol.~2, no.~1, p. e2024JH000542, 2025, e2024JH000542 2024JH000542. [Online]. Available: \url{https://agupubs.onlinelibrary.wiley.com/doi/abs/10.1029/2024JH000542}
\BIBentrySTDinterwordspacing

\bibitem{https://doi.org/10.1029/2020JB020549}
\BIBentryALTinterwordspacing
N.~Wang, H.~Chang, and D.~Zhang, ``Deep-learning-based inverse modeling approaches: A subsurface flow example,'' \emph{Journal of Geophysical Research: Solid Earth}, vol. 126, no.~2, p. e2020JB020549, 2021, e2020JB020549 2020JB020549. [Online]. Available: \url{https://agupubs.onlinelibrary.wiley.com/doi/abs/10.1029/2020JB020549}
\BIBentrySTDinterwordspacing

\bibitem{https://doi.org/10.1029/2019WR026731}
\BIBentryALTinterwordspacing
A.~M. Tartakovsky, C.~O. Marrero, P.~Perdikaris, G.~D. Tartakovsky, and D.~Barajas-Solano, ``Physics-informed deep neural networks for learning parameters and constitutive relationships in subsurface flow problems,'' \emph{Water Resources Research}, vol.~56, no.~5, p. e2019WR026731, 2020, e2019WR026731 10.1029/2019WR026731. [Online]. Available: \url{https://agupubs.onlinelibrary.wiley.com/doi/abs/10.1029/2019WR026731}
\BIBentrySTDinterwordspacing

\bibitem{gmd-16-7375-2023}
\BIBentryALTinterwordspacing
D.~Degen, D.~Caviedes~Voulli\`eme, S.~Buiter, H.-J. Hendricks~Franssen, H.~Vereecken, A.~Gonz\'alez-Nicol\'as, and F.~Wellmann, ``Perspectives of physics-based machine learning strategies for geoscientific applications governed by partial differential equations,'' \emph{Geoscientific Model Development}, vol.~16, no.~24, pp. 7375--7409, 2023. [Online]. Available: \url{https://gmd.copernicus.org/articles/16/7375/2023/}
\BIBentrySTDinterwordspacing

\bibitem{RAISSI2019686}
\BIBentryALTinterwordspacing
M.~Raissi, P.~Perdikaris, and G.~Karniadakis, ``Physics-informed neural networks: A deep learning framework for solving forward and inverse problems involving nonlinear partial differential equations,'' \emph{Journal of Computational Physics}, vol. 378, pp. 686--707, 2019. [Online]. Available: \url{https://www.sciencedirect.com/science/article/pii/S0021999118307125}
\BIBentrySTDinterwordspacing

\bibitem{cite-key}
\BIBentryALTinterwordspacing
G.~E. Karniadakis, I.~G. Kevrekidis, L.~Lu, P.~Perdikaris, S.~Wang, and L.~Yang, ``Physics-informed machine learning,'' \emph{Nature Reviews Physics}, vol.~3, no.~6, pp. 422--440, 2021. [Online]. Available: \url{https://doi.org/10.1038/s42254-021-00314-5}
\BIBentrySTDinterwordspacing

\bibitem{Zahra}
\BIBentryALTinterwordspacing
Z.~Kadkhodaie, F.~Guth, E.~P. Simoncelli, and S.~Mallat, ``Generalization in diffusion models arises from geometry-adaptive harmonic representations,'' 2024. [Online]. Available: \url{https://arxiv.org/abs/2310.02557}
\BIBentrySTDinterwordspacing

\bibitem{22}
D.~Ulyanov, A.~Vedaldi, and V.~Lempitsky, ``\BIBforeignlanguage{lv}{Deep image prior},'' \emph{\BIBforeignlanguage{lv}{\textit{Int J Comput Vis}}}, vol. 128, no.~7, pp. 1867–1888,, 2020.

\bibitem{ref49}
\BIBentryALTinterwordspacing
N.~Rahaman, ``\BIBforeignlanguage{en}{On the spectral bias of neural networks},'' 2019, arXiv,. [Online]. Available: \url{http://arxiv.org/abs/1806.08734}
\BIBentrySTDinterwordspacing

\bibitem{10.1145/3571070}
\BIBentryALTinterwordspacing
G.~Vardi, ``On the implicit bias in deep-learning algorithms,'' \emph{Commun. ACM}, vol.~66, no.~6, p. 86–93, may 2023. [Online]. Available: \url{https://doi.org/10.1145/3571070}
\BIBentrySTDinterwordspacing

\bibitem{frei2022implicitbiasleakyrelu}
\BIBentryALTinterwordspacing
S.~Frei, G.~Vardi, P.~L. Bartlett, N.~Srebro, and W.~Hu, ``Implicit bias in leaky relu networks trained on high-dimensional data,'' 2022. [Online]. Available: \url{https://arxiv.org/abs/2210.07082}
\BIBentrySTDinterwordspacing

\bibitem{allenzhu2019convergencetheorydeeplearning}
\BIBentryALTinterwordspacing
Z.~Allen-Zhu, Y.~Li, and Z.~Song, ``A convergence theory for deep learning via over-parameterization,'' 2019. [Online]. Available: \url{https://arxiv.org/abs/1811.03962}
\BIBentrySTDinterwordspacing

\bibitem{arora2018optimizationdeepnetworksimplicit}
\BIBentryALTinterwordspacing
S.~Arora, N.~Cohen, and E.~Hazan, ``On the optimization of deep networks: Implicit acceleration by overparameterization,'' 2018. [Online]. Available: \url{https://arxiv.org/abs/1802.06509}
\BIBentrySTDinterwordspacing

\bibitem{wilson2022bayesiandeeplearningprobabilistic}
\BIBentryALTinterwordspacing
A.~G. Wilson and P.~Izmailov, ``Bayesian deep learning and a probabilistic perspective of generalization,'' 2022. [Online]. Available: \url{https://arxiv.org/abs/2002.08791}
\BIBentrySTDinterwordspacing

\bibitem{griffiths2023bayesageintelligentmachines}
\BIBentryALTinterwordspacing
T.~L. Griffiths, J.-Q. Zhu, E.~Grant, and R.~T. McCoy, ``Bayes in the age of intelligent machines,'' 2023. [Online]. Available: \url{https://arxiv.org/abs/2311.10206}
\BIBentrySTDinterwordspacing

\bibitem{Physics_of_LLMS}
Z.~{Allen-Zhu}, ``{ICML 2024 Tutorial: Physics of Language Models},'' July 2024, project page: \url{https://physics.allen-zhu.com/}.

\bibitem{kaplan2020scalinglawsneurallanguage}
\BIBentryALTinterwordspacing
J.~Kaplan, S.~McCandlish, T.~Henighan, T.~B. Brown, B.~Chess, R.~Child, S.~Gray, A.~Radford, J.~Wu, and D.~Amodei, ``Scaling laws for neural language models,'' 2020. [Online]. Available: \url{https://arxiv.org/abs/2001.08361}
\BIBentrySTDinterwordspacing

\bibitem{ref18}
\BIBentryALTinterwordspacing
W.~Zhu, K.~Xu, E.~Darve, B.~Biondi, and G.~Beroza, ``\BIBforeignlanguage{en}{Integrating deep neural networks with full-waveform inversion: Reparametrization, regularization, and uncertainty quantification},'' 2021, arXiv,. [Online]. Available: \url{http://arxiv.org/abs/2012.11149}
\BIBentrySTDinterwordspacing

\bibitem{27}
A.~Xu and L.~J. Heagy, ``A test-time learning approach to reparameterize the geophysical inverse problem with a convolutional neural network,'' \emph{IEEE Transactions on Geoscience and Remote Sensing}, vol.~62, pp. 1--12, 2024.

\bibitem{sharma_inversion_2015}
S.~Sharma and G.~K. Verma, ``\BIBforeignlanguage{en}{Inversion of {Electrical} {Resistivity} {Data}: {A} {Review}},'' vol.~9, no.~4, 2015.

\bibitem{wei_quantifying_2022}
\BIBentryALTinterwordspacing
X.~Wei, J.~Sun, and M.~K. Sen, ``Quantifying uncertainty of salt body shapes recovered from gravity data using trans-dimensional {Markov} chain {Monte} {Carlo} sampling,'' \emph{Geophysical Journal International}, vol. 232, no.~3, pp. 1957--1978, Oct. 2022, \_eprint: https://academic.oup.com/gji/article-pdf/232/3/1957/47030191/ggac430.pdf. [Online]. Available: \url{https://doi.org/10.1093/gji/ggac430}
\BIBentrySTDinterwordspacing

\bibitem{melo_geophysical_2017}
\BIBentryALTinterwordspacing
A.~T. Melo, J.~Sun, and Y.~Li, ``Geophysical inversions applied to {3D} geology characterization of an iron oxide copper-gold deposit in {Brazil},'' \emph{Geophysics}, vol.~82, no.~5, pp. K1--K13, Jul. 2017, \_eprint: https://pubs.geoscienceworld.org/seg/geophysics/article-pdf/82/5/K1/4073041/geo-2016-0490.1.pdf. [Online]. Available: \url{https://doi.org/10.1190/geo2016-0490.1}
\BIBentrySTDinterwordspacing

\bibitem{SimPEG}
R.~Cockett, L.~Heagy, and D.~Oldenburg, ``\BIBforeignlanguage{en}{Pixels and their neighbors: Finite volume},'' \emph{\BIBforeignlanguage{en}{\textit{The Leading Edge}}}, vol.~35, no.~8, pp. 703–706,, 2016.

\bibitem{astic_framework_2019}
\BIBentryALTinterwordspacing
T.~Astic and D.~W. Oldenburg, ``\BIBforeignlanguage{en}{A framework for petrophysically and geologically guided geophysical inversion using a dynamic {Gaussian} mixture model prior},'' \emph{\BIBforeignlanguage{en}{Geophysical Journal International}}, vol. 219, no.~3, pp. 1989--2012, Dec. 2019. [Online]. Available: \url{https://academic.oup.com/gji/article/219/3/1989/5556947}
\BIBentrySTDinterwordspacing

\bibitem{ref8}
E.~Laloy, R.~Hérault, J.~Lee, D.~Jacques, and N.~Linde, ``\BIBforeignlanguage{en}{Inversion using a new low-dimensional representation of complex binary geological media based on a deep neural network},'' in \emph{\BIBforeignlanguage{en}{\textit{Advances in Water Resources}}}, vol. 110, pp. 387–405,.

\bibitem{ref11}
J.~Lopez‐Alvis, F.~Nguyen, M.~Looms, and T.~Hermans, ``\BIBforeignlanguage{en}{Geophysical inversion using a variational autoencoder to model an assembled spatial prior uncertainty},'' \emph{\BIBforeignlanguage{en}{\textit{JGR Solid Earth}}}, vol. 127, no.~3, pp. 2021 022\,581,.

\bibitem{rahaman2019spectralbiasneuralnetworks}
\BIBentryALTinterwordspacing
N.~Rahaman, A.~Baratin, D.~Arpit, F.~Draxler, M.~Lin, F.~A. Hamprecht, Y.~Bengio, and A.~Courville, ``On the spectral bias of neural networks,'' 2019. [Online]. Available: \url{https://arxiv.org/abs/1806.08734}
\BIBentrySTDinterwordspacing

\bibitem{sun_coil_2021}
\BIBentryALTinterwordspacing
Y.~Sun, J.~Liu, M.~Xie, B.~Wohlberg, and U.~Kamilov, ``\BIBforeignlanguage{en}{{CoIL}: {Coordinate}-{Based} {Internal} {Learning} for {Tomographic} {Imaging}},'' \emph{\BIBforeignlanguage{en}{IEEE Trans. Comput. Imaging}}, vol.~7, pp. 1400--1412, 2021. [Online]. Available: \url{https://ieeexplore.ieee.org/document/9606601/}
\BIBentrySTDinterwordspacing

\bibitem{kingma2017adammethodstochasticoptimization}
\BIBentryALTinterwordspacing
D.~P. Kingma and J.~Ba, ``Adam: A method for stochastic optimization,'' 2017. [Online]. Available: \url{https://arxiv.org/abs/1412.6980}
\BIBentrySTDinterwordspacing

\bibitem{tan_lim_2019}
H.~H. Tan and K.~H. Lim, ``Review of second-order optimization techniques in artificial neural networks backpropagation,'' \emph{IOP Conference Series: Materials Science and Engineering}, vol. 495, p. 012003, Jun 2019.

\bibitem{he2015delvingdeeprectifierssurpassing}
\BIBentryALTinterwordspacing
K.~He, X.~Zhang, S.~Ren, and J.~Sun, ``Delving deep into rectifiers: Surpassing human-level performance on imagenet classification,'' 2015. [Online]. Available: \url{https://arxiv.org/abs/1502.01852}
\BIBentrySTDinterwordspacing

\bibitem{gmd-15-3161-2022}
\BIBentryALTinterwordspacing
S.~M\"uller, L.~Sch\"uler, A.~Zech, and F.~He{\ss}e, ``\texttt{GSTools} v1.3: a toolbox for geostatistical modelling in python,'' \emph{Geoscientific Model Development}, vol.~15, no.~7, pp. 3161--3182, 2022. [Online]. Available: \url{https://gmd.copernicus.org/articles/15/3161/2022/}
\BIBentrySTDinterwordspacing

\bibitem{kayhan2020translationinvariancecnnsconvolutional}
\BIBentryALTinterwordspacing
O.~S. Kayhan and J.~C. van Gemert, ``On translation invariance in cnns: Convolutional layers can exploit absolute spatial location,'' 2020. [Online]. Available: \url{https://arxiv.org/abs/2003.07064}
\BIBentrySTDinterwordspacing

\bibitem{martel2021acornadaptivecoordinatenetworks}
\BIBentryALTinterwordspacing
J.~N.~P. Martel, D.~B. Lindell, C.~Z. Lin, E.~R. Chan, M.~Monteiro, and G.~Wetzstein, ``Acorn: Adaptive coordinate networks for neural scene representation,'' 2021. [Online]. Available: \url{https://arxiv.org/abs/2105.02788}
\BIBentrySTDinterwordspacing

\bibitem{SIREN}
\BIBentryALTinterwordspacing
V.~Sitzmann, J.~N.~P. Martel, A.~W. Bergman, D.~B. Lindell, and G.~Wetzstein, ``Implicit neural representations with periodic activation functions,'' 2020. [Online]. Available: \url{https://arxiv.org/abs/2006.09661}
\BIBentrySTDinterwordspacing

\end{thebibliography}
\bibliographystyle{IEEEtran}



\vspace{-33pt}

\begin{IEEEbiography}[{\includegraphics[width=1in,height=1.25in,clip,keepaspectratio]{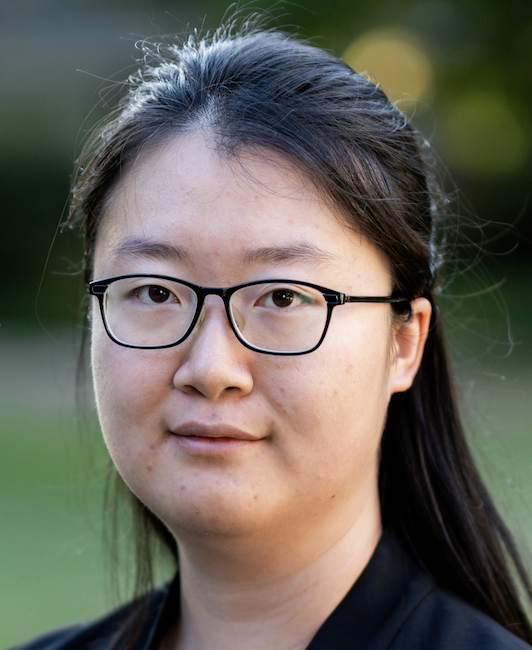}}]{Anran Xu}
received the Honours Bachelor of Science degree in Mathematics $\&$ Its Applications Specialist (Physical Science) and Physics Major from the University of Toronto, Toronto, ON, Canada, in 2022. She received a Master of Science in Geophysics from the University of British Columbia (UBC), Vancouver, BC, Canada, in 2024. She is currently pursuing a PhD in Geophysics under the supervision of Prof. Lindsey Heagy at the Geophysical Inversion Facility from UBC (UBC GIF). 
\par
Her research interests include AI for physics and applications of machine learning models to inverse problems. 
\end{IEEEbiography}
\begin{IEEEbiography}
[{\includegraphics[width=1in,height=1.25in,clip,keepaspectratio]{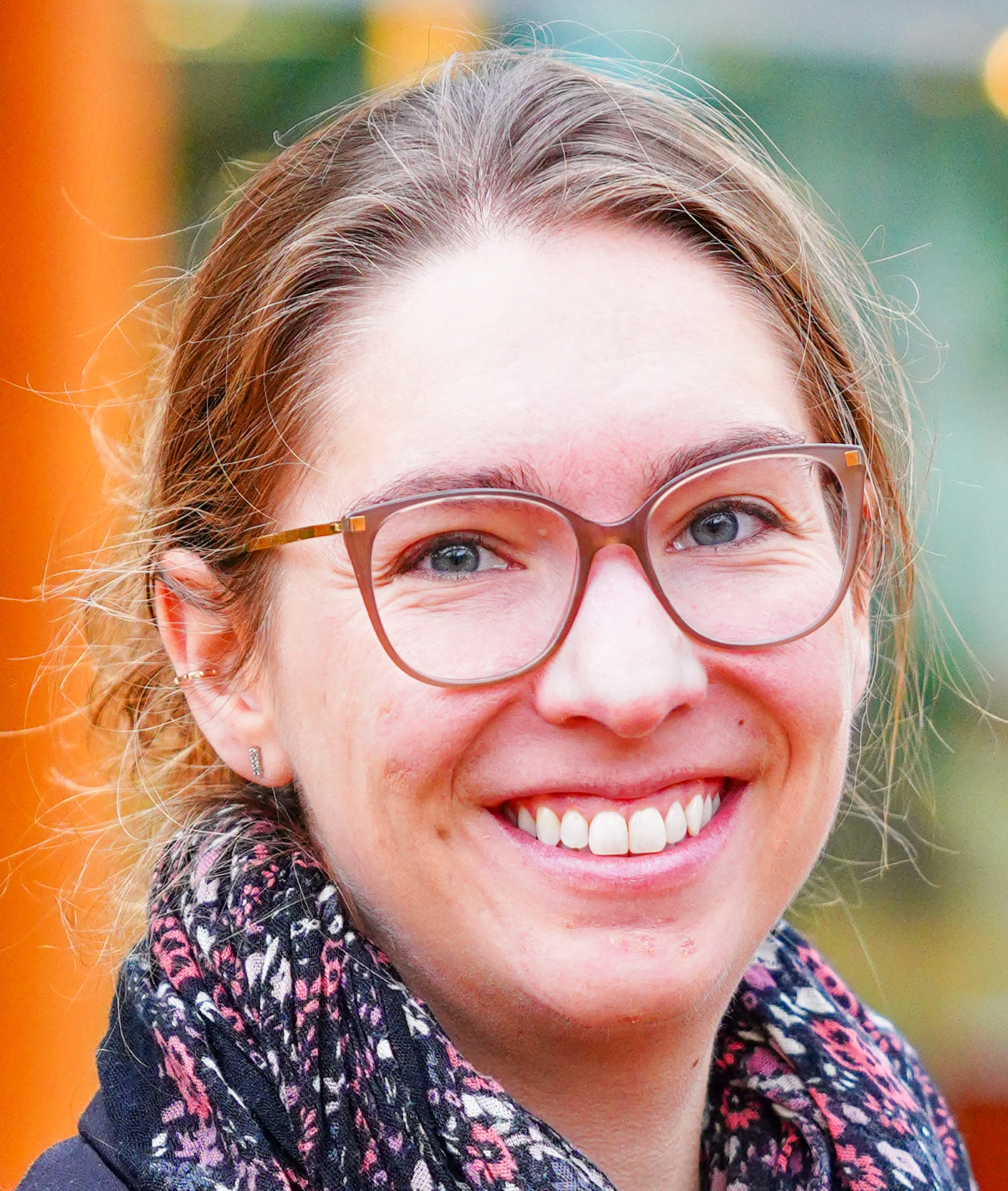}}]{Lindsey J. Heagy} is an Assistant Professor in the Department of Earth, Ocean and Atmospheric Sciences and Director of the Geophysical Inversion Facility at UBC. She completed her BSc in geophysics at the University of Alberta in 2012 and her PhD at UBC in 2018. Prior to her current position, she was a Postdoctoral researcher in the Statistics Department at UC Berkeley. 
\par
Her research combines computational methods in numerical simulations, inversions, and machine learning to characterize the geophysical subsurface.

\end{IEEEbiography}

\vfill

\end{document}